\pdfoutput=1
\documentclass{article} 
\usepackage{iclr2027_conference,times}

\usepackage{amsmath,amsfonts,bm}

\def\eqref#1{equation~\ref{#1}}

\def\1{\bm{1}}

\DeclareMathAlphabet{\mathsfit}{\encodingdefault}{\sfdefault}{m}{sl}
\SetMathAlphabet{\mathsfit}{bold}{\encodingdefault}{\sfdefault}{bx}{n}

\usepackage{hyperref}
\usepackage{url}
\usepackage{booktabs}       
\usepackage{amsfonts}       
\usepackage{nicefrac}       
\usepackage{microtype}      
\usepackage{xcolor}         
\usepackage{amsmath}
\usepackage{amssymb}
\usepackage{algorithm}
\usepackage{algpseudocode}

\usepackage{booktabs}
\usepackage{multirow}
\usepackage{subcaption}
\usepackage{graphicx}
\usepackage{tabularx}
\usepackage{array}
\usepackage{subcaption}

\newcolumntype{L}[1]{%
  >{\raggedright\arraybackslash}p{#1}%
}

\usepackage{listings}

\lstdefinestyle{prompt}{
    basicstyle=\ttfamily\scriptsize,
    breaklines=true,
    breakatwhitespace=false,
    columns=fullflexible,
    keepspaces=true,
    frame=single,
    showstringspaces=false
}
\usepackage[section]{placeins}
\usepackage{float}

\title{MotiveMob: Motivation as Semantic Action for Closed-Loop Human Mobility Generation}

\author{
\textbf{Mengkun Gao}$^{1}$,\;
\textbf{Zengqing Wu}$^{1}$,\;
\textbf{Renhe Jiang}$^{2}$,\;
\textbf{Jiawei Wang}$^{3}$,\;
\textbf{Yusong Wang}$^{4}$ \\
\textbf{Chuang Yang}$^{2}$,\;
\textbf{Shuyuan Zheng}$^{1}$,\;
\textbf{Makoto Onizuka}$^{1}$,\;
\textbf{Chuan Xiao}$^{1\dagger}$ \\
$^{1}$The University of Osaka,\;
$^{2}$The University of Tokyo, \\
$^{3}$The Hong Kong University of Science and Technology,\;
$^{4}$Institute of Science Tokyo \\
\texttt{mengkun.gao@ist.osaka-u.ac.jp},\;
\texttt{wuzengqing@outlook.com} \\
\texttt{jiangrh@csis.u-tokyo.ac.jp},\;
\texttt{wangyi@lr.pi.titech.ac.jp} \\
\texttt{chuang.yang@csis.u-tokyo.ac.jp} \\
\texttt{\{zheng,onizuka,chuanx\}@ist.osaka-u.ac.jp}
}
\iclrfinalcopy
\begin{document}

\maketitle

\begin{abstract}
Human mobility generation, as an important task in urban system research, synthesizes trajectory data that can be used for urban planning, transportation management, etc. Human mobility can be characterized as a ``why--where--when'' decision process: people form an intention to move and then determine where and when the corresponding activity will take place. Trajectory generation under user-level and temporal distribution shifts may benefit from explicitly modeling this decision structure. However, many existing human mobility generation methods either represent behavioral intent at a coarse granularity, such as a daily plan or a trajectory-level description, or directly predict future locations without explicitly reasoning about a possible motivation for each movement step. We introduce \textbf{MotiveMob}, a motivation-driven autoregressive framework for human mobility generation that first forms a hypothesis about why the next movement may occur and then jointly generates where and when it may occur. At each step, a motivation predictor conditions on the current mobility state, a long-term behavioral report, and the mobility history to infer a plausible motivation or determine whether the trajectory should terminate. Given the hypothesized motivation, a state predictor grounds it in a candidate next location and arrival time. Before being fed back as the current mobility state for the next ``why--where--when'' decision, the candidate is subject to speed-feasibility and repetition checks. We evaluate MotiveMob under distribution shifts involving unseen users and unseen temporal periods, including seasonal changes and the substantial behavioral disruption caused by the COVID-19 pandemic. Experiments show that MotiveMob consistently achieves better distributional fidelity than competitive pretraining-based and prompting-based methods under user-level and temporal distribution shifts, demonstrating robust generalization to out-of-distribution mobility patterns.
\end{abstract}

\section{Introduction}

Human mobility trajectory generation aims to synthesize realistic sequences of visits, including the locations individuals visit and the corresponding timestamps \citep{luca2021survey}. High-quality synthetic trajectories support a wide range of applications, including urban planning, transportation management, and epidemic modeling \citep{zhu2023synmob, lopez2018microscopic, chang2021mobility}. Recent advances in large language models (LLMs) have inspired new approaches to this task. Some methods fine-tune LLMs for next-point-of-interest (POI) prediction, while others prompt LLMs to generate daily mobility plans from semantic descriptions and historical observations \citep{li2024large,wang2025generative,wang2024large,wang2026ellmob}.

From a behavioral perspective, human mobility can be viewed as a motivation-first decision process: people form an intention to move and then determine both where to go and when to arrive. This factorization is particularly important for human mobility generation because similar mobility states may lead to different next movements depending on the latent motivation. However, many existing methods either predict future locations directly from observed trajectories or represent intent only at the coarse granularity of a daily plan, without explicitly modeling the step-level motivation associated with each transition. As a result, they may conflate behaviorally distinct transitions and struggle to maintain consistent activity and temporal patterns over extended generation horizons. This limitation becomes particularly consequential under user-level and temporal distribution shifts, as mobility patterns vary across individuals and evolve with seasonal changes and disruptions such as the COVID-19 pandemic. Generalization under these shifts may benefit from representations of users' long-term behavioral patterns and recent activities, together with a transition model that maps hypothesized motivations to plausible next states. Although prior work suggests that LLMs can encode latent environment dynamics and support implicit world modeling~\citep{abdou2021can,li2022emergent,zhang2025agent}, mobility data record only realized transitions and provide no direct evidence of what would have occurred under alternative motivations, making such dynamics difficult to learn from observed trajectories alone.

Seeing this limitation in existing studies, we introduce \textbf{MotiveMob}, an LLM-based autoregressive framework that generates daily mobility trajectories through step-wise ``why--where--when'' decisions. At each step, a motivation predictor $P_M$ conditions on the current mobility state, a long-term behavioral report $\mathcal{R}$, and a recent mobility history $\mathcal{H}$ to predict the motivation for the next movement or determine whether the trajectory should terminate. A state predictor $P_S$ then conditions on the same user context and treats the predicted motivation as a semantic action, grounding it in a feasible next location and arrival time. To train $P_S$ beyond observed transitions, we construct pseudo alternative transitions by pairing observed states with alternative motivation labels, generating candidate next states with a lightweight predictor, and using an LLM-based judge to select the most plausible candidate. The observed and alternative transitions are then used to train $P_S$ as a motivation-conditioned transition model, which serves as the dynamics component of an implicit world model (IWM). During inference, $P_M$ and $P_S$ operate alternately. Speed-infeasible candidates are regenerated, whereas repeated states terminate the rollout. Each accepted state then becomes the current mobility state for the next decision.

We evaluate MotiveMob under both user-level and temporal distribution shifts, covering seen and unseen users across an observed period, a regular unseen period, and a COVID-19 period characterized by substantial changes in mobility behavior. Across these settings, MotiveMob consistently achieves better distributional fidelity than competitive pretraining-based and prompting-based methods. The advantage is particularly evident in time--activity distribution fidelity: averaged across the three temporal periods, MotiveMob reduces the corresponding divergence by 10.7\% for seen users and 18.3\% for unseen users relative to the best-performing baseline in each setting. Compared with the variant without $P_M$, the corresponding reductions are 21.6\% and 29.1\%, respectively, supporting the benefit of reasoning about why a movement may occur before generating where and when it occurs. In addition, the evaluation analyzes the roles and metric-specific trade-offs of implicit world modeling, user-specific context in trajectory generation, and inference-time rollout validation.

Our contributions are summarized as follows:

\begin{itemize}

\item We formulate daily mobility trajectory generation as a step-level, motivation-first ``why--where--when'' process and introduce \textbf{MotiveMob}, which separates potential motivation prediction from the joint generation of the next location and arrival time.

\item We develop an offline alternative transition augmentation procedure for training the state predictor as an IWM. A lightweight candidate generator proposes outcomes for hypothetical motivations, while an LLM-based judge filters plausible transitions, enabling the model to learn beyond observed state--motivation pairs without requiring online environment interaction.

\item We conduct a systematic evaluation under user-level and temporal distribution shifts, covering unseen users, regular seasonal variation, and pandemic-induced behavioral disruption. MotiveMob consistently achieves better distributional fidelity than competitive pretraining-based and prompting-based baselines.

\end{itemize}
\section{Related Work}

\subsection{Human Mobility Modeling}

Human mobility modeling encompasses next-location prediction and multi-step trajectory generation \citep{luca2021survey,kong2023mobility}. Next-location prediction infers a user's next destination from historical visits and contextual information. Early approaches used probabilistic transition models \citep{gambs2012next,lu2013approaching}, followed by recurrent and attention-based architectures that capture long-term dependencies and individual preferences \citep{feng2018deepmove,luo2021stan}. Recent LLM-based methods employ time-aware prompting or zero-shot candidate ranking \citep{wang2023would,feng2024move}, while others adapt LLMs through task-specific fine-tuning \citep{li2024large,wang2025generative}; Mobility-LLM further learns intention- and preference-aware representations for predictive check-in tasks \citep{gong2024mobility}. These methods primarily target one-step or related predictive objectives. Trajectory generation instead synthesizes complete location--time sequences and aims to reproduce population-level spatiotemporal regularities. Existing generators employ variational autoencoders \citep{huang2019variational}, adversarial imitation learning \citep{zhang2020trajgail,choi2021trajgail}, and diffusion models \citep{zhu2023difftraj,chu2024simulating,long2026dynamic}. More recent LLM-based methods formulate trajectories as conditional sequences or agent-generated plans \citep{ohara2024generating,zhang2024mobglm,li2025geo,wang2026ellmob,Li2026trajgenagent}, as summarized in a recent survey \citep{gao2026llms}. Particularly relevant activity-centric approaches generate activity--time chains before assigning spatial locations \citep{liao2026deep}, while retrieval-augmented methods synthesize household-coordinated activity chains from aggregate statistics and sociodemographic profiles \citep{liu2025human}. LLMob also infers a daily-level motivation to guide whole-day trajectory generation \citep{wang2024large}. However, existing activity-centric generators typically introduce semantic conditioning at the level of whole-day motivations or activity chains, rather than explicitly modeling the motivation associated with each individual transition. Moreover, how such models generalize to unseen users and future or disrupted periods also remains underexplored.

\subsection{World Modeling with LLMs}
In reinforcement learning (RL), it is often desirable for an agent to have knowledge of the environment dynamics, enabling it to model state transitions and predict future states and rewards, as in model-based RL \citep{luo2024survey}. Traditionally, such knowledge is acquired by learning a world model from observed state transitions, ranging from simple lookup-based models \citep{sutton1991dyna} to deep generative models \citep{ha2018recurrent}. With the emergence of LLMs, agents have begun to employ LLMs as world models to support planning and decision making \citep{gu2024your,feng2025web}. However, most prior approaches treat the world model as an explicit and separate simulator, echoing classical control pipelines that rely on planning over imagined rollouts. In contrast, \citet{zhang2025agent} propose an early experience paradigm in which the policy is trained to predict its own future states from an IWM, thereby internalizing environment dynamics without constructing a standalone simulator. Learning action-conditioned dynamics, however, requires transition data that cover the consequences of different actions under comparable states. Human mobility records contain only real transitions and do not reveal what might have occurred under alternative motivations. This lack of counterfactual observations, together with the absence of an interactive mobility environment, presents a key challenge for learning IWMs in trajectory generation.
\section{Preliminaries}


\subsection{Daily Trajectory Generation}
For a user $u$ and a target date $d$, we represent a daily trajectory as an ordered sequence of location--time pairs,
\begin{equation}
\tau_{u,d}
=
\left\langle
(l_0,t_0),(l_1,t_1),\ldots,(l_{n_{u,d}-1},t_{n_{u,d}-1})
\right\rangle,
\end{equation}
where $l_i$ denotes the POI identifier of the $i$-th check-in and $t_i$ denotes its arrival timestamp, $(l_i,t_i)$ denotes the user's $i$-th check-in, and $n_{u,d}$ denotes the number of check-ins in the trajectory. Given the user's long-term behavioral report $\mathcal{R}_u$, recent mobility history $\mathcal{H}_{u,d}$, and target date $d$, our objective is to generate $\tau_{u,d}$, describing the sequence of locations visited by the user and their corresponding visit times during the target mobility day. The recent mobility history $H_{u,d}$ always precedes the target date, whereas $R_u$ is constructed from a query-disjoint support set from the seen period. The construction and textual representations of $\mathcal{R}_u$ and $\mathcal{H}_{u,d}$ are detailed in Appendix~\ref{app:user-context}.



\subsection{Mobility Dynamics as a World Model}
From a world model perspective, each check-in is represented as a state $s_i=(l_i,t_i)\in\mathcal{S}$, and each transition from $s_i$ to $s_{i+1}$ is conditioned on a semantic action $m_i\in\mathcal{M}$. We interpret $m_i$ as a transition-level latent variable representing a plausible motivation for the user's next movement, such as work, shopping, or healthcare. Because psychological intent is not directly observed in mobility traces, we weakly supervise $m_i$ using labels derived from destination categories, which serve as operational proxies rather than ground-truth annotations of human intent; the full taxonomy is provided in Appendix~\ref{app:motivation-taxonomy}.For a user $u$ and a target date $d$, the corresponding mobility dynamics can be expressed as
\begin{equation}
s_{i+1}
\sim
\mathcal{T}
\left(
\cdot
\mid
s_i,m_i,\mathcal{R}_u,\mathcal{H}_{u,d},d
\right).
\end{equation}
which characterizes the distribution of the next location and arrival time resulting from a motivation under the user's behavioral context.

\section{Methodology}
As illustrated in Figure~\ref{fig:motivemob-framework}, we introduce \textbf{MotiveMob}, an LLM-based framework that generates daily mobility trajectories by repeatedly making ``why--where--when'' decisions. Unlike existing approaches that reason globally about an entire trajectory or daily plan \citep{wang2024large,wang2026ellmob}, MotiveMob decomposes mobility generation into a sequence of motivation-conditioned check-in transitions. At each step, the motivation predictor $P_M$ conditions on the current mobility state, the user's long-term behavioral report $\mathcal{R}_u$, and recent mobility history $\mathcal{H}_{u,d}$ to infer the latent motivation for the next move or determine whether the trajectory should terminate. The state predictor $P_S$ then treats the inferred motivation as a semantic action and grounds it in a candidate next location and arrival time, after which the candidate transition undergoes inference-time validation. The two predictors are trained separately: $P_S$ learns from observed and alternative transitions, whereas $P_M$ is fine-tuned with step-level motivation labels. During inference, each accepted state becomes the current mobility state for the next decision.

\DeclareRobustCommand{\rollouticon}{%
  \raisebox{-0.15\height}{%
    \includegraphics[height=1.05em]{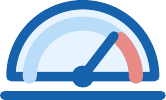}%
  }%
}

\begin{figure*}[t]
    \centering
    \includegraphics[width=\textwidth]{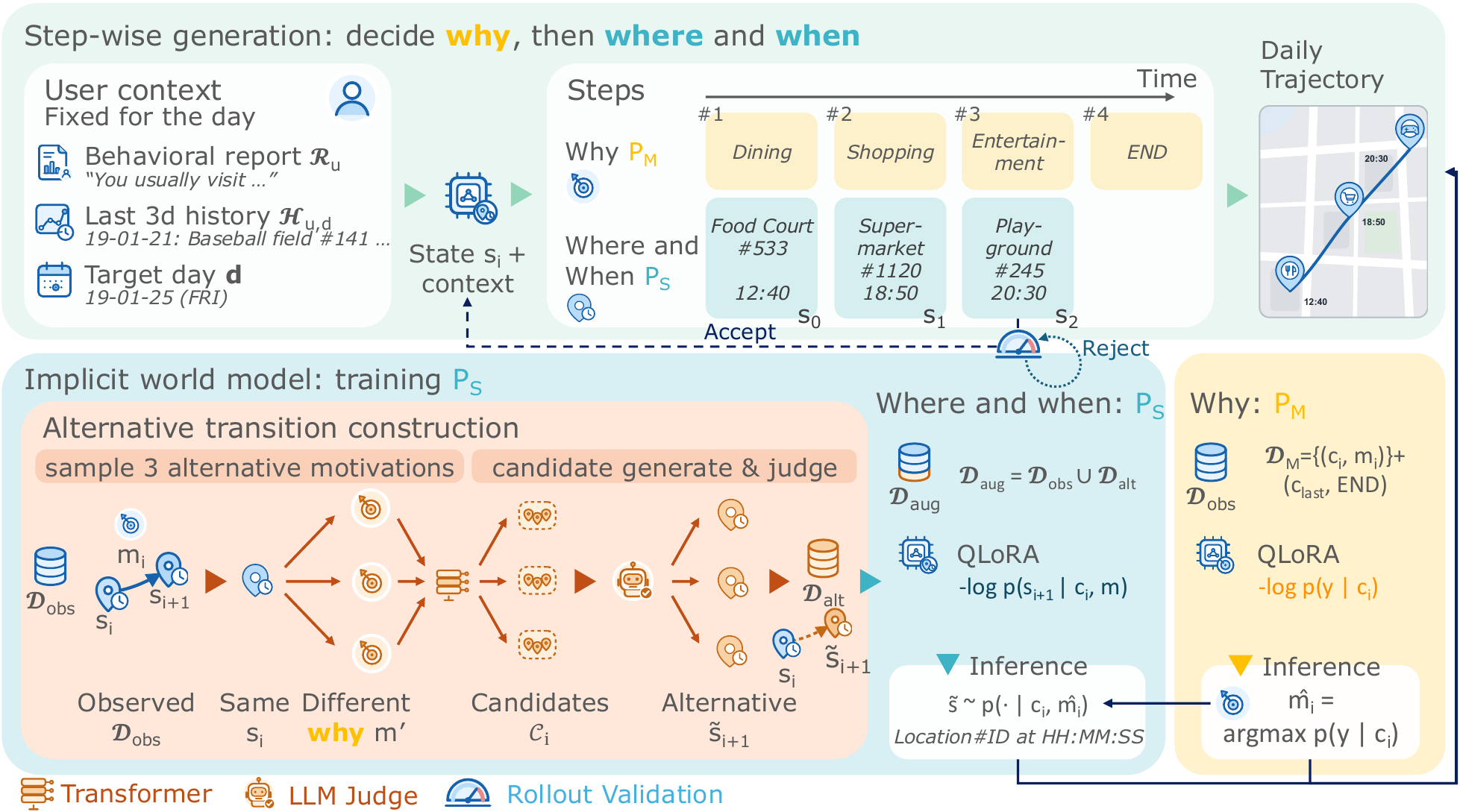}
    \caption[Overview of the MotiveMob framework.]{
    Overview of the MotiveMob framework.
    Given the user context, MotiveMob generates a daily trajectory step by
    step, first predicting \emph{why} ($P_M$) and then \emph{where and when}
    ($P_S$). The implicit world model $P_S$ is trained using observed and alternative transitions. The \rollouticon\ marker denotes \emph{rollout validation}, which performs speed-feasibility checking and repetition-based termination, as described in Section~\ref{subsec:motivation-guided-generation}.
    }
    \label{fig:motivemob-framework}
\end{figure*}

\subsection{State Predictor}
\subsubsection{Alternative Transition Construction}
Because human mobility records contain only real transitions and do not reveal the outcomes of alternative motivations, we augment the observed data with model-synthesized alternative transitions constructed through a two-stage pipeline. Let $\mathcal{D}_{\mathrm{obs}} =\{(s_i,m_i,s_{i+1})\}_{i=1}^{N}$ denote the set of observed transitions. For each observed transition in $\mathcal{D}_{\mathrm{obs}}$, we sample three alternative motivations $\{m_i^j\}_{j=1}^{3}\subset\mathcal{M}\setminus\{m_i\}$, each representing a motivation that the user could have adopted from the same state $s_i$. In the first stage, a lightweight Transformer-based candidate generator \citep{vaswani2017attention} conditions on the trajectory history, the current state $s_i$, and an alternative motivation $m_i^j$ to produce a set of candidate next states $\mathcal{C}_i^j$. In the second stage, an LLM-based judge evaluates these candidates using the user's historical information $\mathcal{H}_i$, the specified motivation $m_i^j$, and their spatial and temporal distances from the current state, denoted by $\boldsymbol{\delta}_i^j$. It then selects the candidate most consistent with the user's behavioral profile, mobility context, and the alternative motivation:
\begin{equation}
\tilde{s}_{i+1}^{j}
=
\operatorname{Judge}_{\mathrm{LLM}}
\left(
\mathcal{C}_i^j,
\mathcal{H}_i,
m_i^j,
\boldsymbol{\delta}_i^j
\right).
\end{equation}

The selected states form a pool of pseudo-labeled alternative transitions,
denoted by $\mathcal{D}_{\mathrm{alt}}$. To balance observed and alternative
supervision during training, we uniformly retain a fraction $\rho$ of this
pool, yielding $\mathcal{D}_{\mathrm{alt}}^{(\rho)} \subseteq \mathcal{D}_{\mathrm{alt}}$. The augmented training set is then defined as $\mathcal{D}_{\mathrm{aug}}= \mathcal{D}_{\mathrm{obs}} \cup \mathcal{D}_{\mathrm{alt}}^{(\rho)}.$

\subsubsection{Implicit World Modeling}
Using the augmented transition set $\mathcal{D}_{\mathrm{aug}}$, we train the state predictor $P_S$ as an IWM of motivation-conditioned mobility dynamics. Let
$c_i=(s_i,\mathcal{R}_u,\mathcal{H}_{u,d},d)$
denote the context available at step $i$, including the current mobility state,
the behavioral report of user $u$, the recent mobility history preceding the
target date, and the target date $d$. Given a context $c_i$ and a motivation
$m$ treated as a semantic action, $P_S$ models the resulting next-state
distribution as
$p_{\theta_S}(s'\mid c_i,m)$.
For training, we associate each transition with its corresponding context and
represent each instance in $\mathcal{D}_{\mathrm{aug}}$ as $(c_i,m,s')$. We
train $P_S$ by minimizing the average negative log-likelihood:
\begin{equation}
\mathcal{L}_S
=
-\frac{1}{|\mathcal{D}_{\mathrm{aug}}|}
\sum_{(c_i,m,s')\in\mathcal{D}_{\mathrm{aug}}}
\log p_{\theta_S}(s'\mid c_i,m).
\end{equation}
This yields an offline, reward-free world modeling objective that combines
factual grounding with broader coverage of motivation-conditioned mobility
dynamics.

\subsection{Motivation Predictor}

The state predictor $P_S$ grounds a given motivation in a next mobility state but does not determine which motivation should drive the next move. We therefore train a dedicated motivation predictor $P_M$ to infer a step-level potential motivation, or predict trajectory termination, from the context $c_i=(s_i,\mathcal{R}_u,\mathcal{H}_i)$. To prevent future information leakage, $c_i$ contains only information available up to step $i$ and excludes the next state $s_{i+1}$ and all subsequent states. Besides, for each observed trajectory, we append a terminal training instance whose label is $\mathtt{END}$. We construct the motivation prediction dataset as
$\mathcal{D}_M = \{(c_i,y_i)\}_{i=1}^{N_M},$
where $y_i=m_i$ for an observed transition and $y_i=\mathtt{END}$ when the trajectory should terminate. We fine-tune $P_M$ by minimizing the average negative log-likelihood:
\begin{equation}
\mathcal{L}_M
=
-\frac{1}{|\mathcal{D}_M|}
\sum_{(c_i,y_i)\in\mathcal{D}_M}
\log p_{\theta_M}(y_i\mid c_i).
\end{equation}
This supervised objective enables $P_M$ to infer the behavioral intent underlying the next move or determine that no further transition should be generated. During generation, a predicted motivation is passed to $P_S$ for state grounding, whereas $\mathtt{END}$ terminates the trajectory.

\subsection{Motivation-Guided Trajectory Generation}
\label{subsec:motivation-guided-generation}

At inference time, $P_M$ and $P_S$ operate sequentially to generate a daily trajectory through step-level``why--where--when'' decisions. Generation begins with a $\mathtt{START}$ token in the textual context. Given this context, $P_M$ predicts the first motivation and $P_S$ generates the initial physical state $s_0$. Speed validation is applied from the subsequent transition onward. At each step $i$, $P_M$ greedily predicts
$\hat{m}_i = \arg\max_{y\in\mathcal{M}\cup\{\mathtt{END}\}}p_{\theta_M}(y\mid c_i).$
If $\hat{m}_i=\mathtt{END}$, generation terminates. Otherwise, $P_S$ conditions on $c_i$ and $\hat{m}_i$ to sample a provisional next state
$\tilde{s}_{i+1} \sim p_{\theta_S}(\,\cdot\mid c_i,\hat{m}_i),$
where $\tilde{s}_{i+1}$ specifies the next location and arrival time. Once accepted, $\tilde{s}_{i+1}$ becomes the current state in $c_{i+1}$, and the same procedure is repeated for the next transition.

\paragraph{Inference-Time Rollout Validation.}
Before a provisional state is added to the trajectory, it is validated for temporal and travel feasibility. Specifically, its arrival time must not precede the current time, and its implied travel speed must not exceed $v_{\max}=50\text{ km/h}$. The selection of this threshold is discussed in Appendix~\ref{app:speed-threshold}. An invalid proposal is discarded and resampled from $P_S$. If no proposal satisfies the constraints within a fixed number of attempts, the last proposal is re-grounded to a feasible location of the same activity category using the user's observed mobility information~\citep{Li2026trajgenagent}. If no such feasible location is available, the final model proposal is retained unchanged. We additionally apply repetition-aware termination once the generated trajectory reaches a minimum length: the rollout terminates if the candidate-augmented location sequence contains a short consecutively repeating pattern. Generation also terminates when the prescribed maximum trajectory length is reached.

Through this iterative process, $P_M$ determines why the next move occurs, while $P_S$ grounds the predicted motivation in where and when it occurs. Additional implementation details are provided in the appendices. Appendix~\ref{app:training-inference} provides implementation details for alternative transition construction, predictor training, and autoregressive trajectory inference; Appendix~\ref{app:prompts} provides the prompts used for training and inference; and Appendix~\ref{app:trajectory-generation} presents pseudocode for inference-time trajectory generation.

\section{Experiments}
\label{sec: exp}
\subsection{Experimental Setup}

\textbf{Dataset.} All experiments are conducted on a human mobility dataset collected in Tokyo through the Twitter and Foursquare APIs. The original dataset consists of individual check-ins with location and timestamp information. We transform consecutive check-ins from the same user into transitions, each comprising the origin and destination states and one of 11 behavioral motivation labels derived from the category of the destination venue. We additionally append an \texttt{END} label after the final check-in to mark trajectory termination. For the seen-user, seen-time setting, we randomly hold out 20\% of each user's trajectories for evaluation and use the remaining 80\% for training.

\textbf{Distribution-Shift Settings.} We evaluate generalization along both user and temporal dimensions. The training set contains trajectories from 25 seen users collected between January and October 2019. For user-level generalization, we evaluate exclusively on an additional 50 users whose trajectories are not included in training. For temporal generalization, we consider two unseen periods: November 2019 to February 2020, representing a regular seasonal shift, and April to October 2020, representing a pandemic-induced mobility shift following the onset of COVID-19 restrictions. Detailed split construction and leakage prevention are provided in Appendix~\ref{app:split_protocol}.


\textbf{Baseline Methods.} We compare MotiveMob with two categories of baselines: (1) pretraining-based models, including LSTM~\citep{hochreiter1997long}, MHSA~\citep{vaswani2017attention}, DeepMove~\citep{feng2018deepmove}, GetNext~\citep{yang2022getnext}, TrajGAIL~\citep{choi2021trajgail}, DiffTraj~\citep{zhu2023difftraj}, GNPR-SID~\citep{wang2025generative}, and LLM4POI~\citep{li2024large}; and (2) prompting-based models, including LLM-Mob~\citep{wang2023would}, LLMMove~\citep{feng2024move}, LLM-ZS~\citep{beneduce2025large}, LLMob~\citep{wang2024large}, and ELLMob~\citep{wang2026ellmob}.

\textbf{Evaluation Metrics.} Following \citet{yuan2022activity} and \citet{wang2024large}, we evaluate generated trajectories using four distributional metrics: \textit{Step Distance (SD)}, the geographic distance between consecutive locations; \textit{Step Interval (SI)}, the elapsed time between consecutive check-ins; \textit{Daily Activity Routine Distribution (DARD)}, the joint distribution of discretized time slot $t$ and activity category $c$; and \textit{Spatiotemporal Visit Distribution (STVD)}, the joint distribution of time and location, represented by $(t,\mathrm{lat},\mathrm{lng})$. For each metric, we compute the Jensen--Shannon divergence (JSD) between the generated and real distributions, with lower values indicating better performance.

\subsection{Results and Analysis}
Tables~\ref{tab:seen-users-temporal} and~\ref{tab:unseen-users-temporal} report the results for seen and unseen users across the three temporal settings. All methods use the same validation and termination rules. MotiveMob achieves lower JSD than all baselines across both user groups and all three temporal settings. Its advantage persists for users excluded from training, indicating that its motivation-conditioned modeling transfers beyond observed user identities. Although temporal shifts generally increase divergence, with the COVID-19 period producing the largest mismatch in most dimensions, MotiveMob maintains its advantage under both regular seasonal and pandemic-induced shifts. In contrast, the baselines exhibit metric-specific strengths but do not perform consistently well across mobility dimensions. To further examine MotiveMob across both urban contexts and backbone choices, Appendix~\ref{app:osaka} reports additional results for Osaka, while Appendix~\ref{app:motivemob-backbone-comparison} evaluates MotiveMob with different LLM backbones.

\begin{table*}[t]
\centering
\caption{Evaluation results for seen and unseen users across the three temporal settings. Panels (a) and (b) report results for seen and unseen users, respectively. In each panel, the upper and lower baseline sections correspond to pretraining-based and prompting-based methods, respectively. All values are JSD; lower is better. The best result in each column of each panel is shown in bold, and the runner-up is underlined. Speed and Rep. denote speed validation and repetition-aware termination, respectively. Best results are in bold, and runner-up results are underlined.}
\label{tab:user-temporal-results}
\setlength{\tabcolsep}{4pt}
\renewcommand{\arraystretch}{1.12}

\begin{subtable}{\textwidth}
\centering
\caption{Seen users}
\label{tab:seen-users-temporal}
\resizebox{\linewidth}{!}{%
\begin{tabular}{l*{12}{c}}
\toprule
\multirow{2}{*}{Method}
& \multicolumn{4}{c}{Seen period (908 trajectories)}
& \multicolumn{4}{c}{Regular unseen period (690 trajectories)}
& \multicolumn{4}{c}{COVID-19 unseen period (392 trajectories)} \\
\cmidrule(lr){2-5}
\cmidrule(lr){6-9}
\cmidrule(l){10-13}
& SD & SI & DARD & STVD
& SD & SI & DARD & STVD
& SD & SI & DARD & STVD \\
\midrule
LSTM
& 0.0915 & 0.1042 & 0.2510 & 0.4079
& 0.1014 & 0.1530 & 0.3046 & 0.5159
& 0.0823 & 0.1600 & 0.3713 & 0.6267 \\
MHSA
& 0.0319 & 0.0507 & 0.1898 & 0.4049
& 0.0393 & 0.0590 & 0.2157 & 0.4753
& 0.0424 & 0.0621 & 0.2638 & 0.6201 \\
DeepMove
& 0.0830 & 0.0990 & 0.2334 & 0.4991
& 0.0593 & 0.1305 & 0.3592 & 0.5331
& 0.0911 & 0.1372 & 0.4313 & 0.6422 \\
GetNext
& 0.0440 & 0.2414 & 0.2126 & 0.4917
& 0.0559 & 0.2387 & 0.2114 & 0.5659
& 0.0773 & 0.2046 & 0.1991 & 0.6613 \\
TrajGAIL
& 0.1079 & 0.1382 & 0.1140 & 0.5910
& 0.1183 & 0.1295 & 0.1360 & 0.6367
& 0.1311 & 0.1363 & 0.2136 & 0.6570 \\
DiffTraj
& 0.2000 & 0.1204 & 0.4777 & 0.6931
& 0.2193 & 0.1125 & 0.5157 & 0.6898
& 0.2595 & 0.1052 & 0.5716 & 0.6903 \\
GNPR-SID
& 0.0171 & 0.0163 & 0.1006 & 0.4167
& 0.0145 & 0.0207 & 0.1103 & 0.4799
& 0.0491 & 0.0310 & 0.1445 & 0.5723 \\
LLM4POI
& 0.0098 & 0.0236 & 0.1090 & 0.3788
& 0.0138 & 0.0340 & 0.1329 & 0.4622
& 0.0452 & 0.0217 & 0.1405 & 0.5750 \\
\midrule
LLMMove
& 0.0839 & 0.1164 & 0.1906 & 0.5637
& 0.0830 & 0.1091 & 0.1901 & 0.5607
& 0.1137 & 0.0937 & 0.2572 & 0.6041 \\
LLM-ZS
& 0.0306 & 0.1034 & 0.2369 & 0.5936
& 0.0441 & 0.1016 & 0.2505 & 0.6014
& 0.0732 & 0.0813 & 0.2817 & 0.6295 \\
LLM-Mob
& 0.0951 & 0.1171 & 0.3242 & 0.5639
& 0.1166 & 0.1135 & 0.3336 & 0.5856
& 0.1328 & 0.0941 & 0.3501 & 0.6201 \\
LLMob
& 0.0919 & 0.2496 & 0.1819 & 0.5461
& 0.0820 & 0.2422 & 0.2353 & 0.5972
& 0.0982 & 0.2610 & 0.2013 & 0.6229 \\
ELLMob
& 0.0501 & 0.1053 & 0.2791 & 0.4598
& 0.0476 & 0.1085 & 0.2863 & 0.4937
& 0.0583 & 0.1245 & 0.3309 & 0.5836 \\
\midrule
\textbf{MotiveMob}
& \textbf{0.0062} & \textbf{0.0100} & \underline{0.0765} & \textbf{0.3658}
& \textbf{0.0102} & \underline{0.0161} & \underline{0.0995} & \textbf{0.4301}
& 0.0312 & \textbf{0.0147} & 0.1377 & \textbf{0.5487} \\
MotiveMob w/o $P_M$
& 0.0106 & 0.0168 & 0.1092 & 0.3757
& 0.0179 & 0.0174 & 0.1300 & 0.4541
& \textbf{0.0253} & 0.0233 & 0.1610 & \underline{0.5530} \\
MotiveMob w/o $\mathcal{R}_u$
& 0.0168 & 0.0219 & 0.0912 & 0.5142
& 0.0185 & 0.0259 & 0.1030 & 0.5583
& 0.0525 & \underline{0.0193} & \textbf{0.1266} & 0.5836 \\
MotiveMob w/o $\mathcal{H}_{u,d}$
& 0.0120 & 0.0291 & 0.1422 & 0.4291
& 0.0142 & 0.0343 & 0.1760 & 0.4887
& 0.0805 & 0.0299 & 0.1875 & 0.5945 \\
MotiveMob w/o Speed
& 0.0276 & \underline{0.0109} & \textbf{0.0735} & 0.3762
& 0.0250 & \textbf{0.0152} & \textbf{0.0951} & \underline{0.4351}
& 0.0878 & 0.0241 & \underline{0.1300} & 0.5548 \\
MotiveMob w/o Rep.
& \underline{0.0074} & 0.0161 & 0.0891 & \underline{0.3749}
& \underline{0.0125} & 0.0235 & 0.1114 & 0.4423
& \underline{0.0281} & 0.0245 & 0.1461 & 0.5578 \\
\bottomrule
\end{tabular}%
}
\end{subtable}

\vspace{6pt}

\begin{subtable}{\textwidth}
\centering
\caption{Unseen users}
\label{tab:unseen-users-temporal}
\resizebox{\linewidth}{!}{%
\begin{tabular}{l*{12}{c}}
\toprule
\multirow{2}{*}{Method}
& \multicolumn{4}{c}{Seen period (1345 trajectories)}
& \multicolumn{4}{c}{Regular unseen period (795 trajectories)}
& \multicolumn{4}{c}{COVID-19 unseen period (541 trajectories)} \\
\cmidrule(lr){2-5}
\cmidrule(lr){6-9}
\cmidrule(l){10-13}
& SD & SI & DARD & STVD
& SD & SI & DARD & STVD
& SD & SI & DARD & STVD \\
\midrule
LSTM
& 0.3607 & 0.3354 & 0.5257 & 0.6193
& 0.3610 & 0.3556 & 0.5601 & 0.6515
& 0.3456 & 0.3554 & 0.5648 & 0.6786 \\
MHSA
& 0.1779 & 0.0733 & 0.3335 & 0.5796
& 0.1510 & 0.0895 & 0.3538 & 0.6122
& 0.1557 & 0.0918 & 0.3639 & 0.6716 \\
DeepMove
& 0.2247 & 0.1621 & 0.2977 & 0.6660
& 0.2517 & 0.1896 & 0.4468 & 0.6883
& 0.2393 & 0.1889 & 0.4370 & 0.6897 \\
GetNext
& 0.0715 & 0.1930 & 0.2303 & 0.6223
& 0.0627 & 0.1818 & 0.2419 & 0.6561
& 0.0760 & 0.1931 & 0.2246 & 0.6703 \\
TrajGAIL
& 0.0999 & 0.0801 & 0.1051 & 0.5941
& 0.1237 & 0.0808 & 0.1273 & 0.6222
& 0.1079 & 0.0866 & 0.1540 & 0.6638 \\
DiffTraj
& 0.2176 & 0.0671 & 0.5589 & 0.6915
& 0.2082 & 0.0708 & 0.5681 & 0.6914
& 0.2327 & 0.0773 & 0.5459 & 0.6921 \\
GNPR-SID
& 0.0270 & 0.0488 & 0.1428 & 0.4762
& 0.0263 & 0.0508 & 0.1514 & 0.5356
& 0.0388 & 0.0543 & 0.1706 & 0.6269 \\
LLM4POI
& 0.0469 & 0.0329 & 0.1152 & 0.5391
& 0.0516 & 0.0339 & 0.1452 & 0.5873
& 0.0553 & 0.0369 & 0.1502 & 0.6386 \\
\midrule
LLMMove
& 0.0914 & 0.1376 & 0.1523 & 0.5303
& 0.0787 & 0.1330 & 0.1612 & 0.5921
& 0.1044 & 0.1557 & 0.1802 & 0.6253 \\
LLM-ZS
& 0.0313 & 0.1049 & 0.2171 & 0.5713
& 0.0373 & 0.1221 & 0.2394 & 0.6212
& 0.0442 & 0.1122 & 0.2449 & 0.6473 \\
LLM-Mob
& 0.0948 & 0.0991 & 0.3200 & 0.5314
& 0.1007 & 0.1015 & 0.3284 & 0.5846
& 0.1196 & 0.1091 & 0.3540 & 0.6271 \\
LLMob
& 0.0555 & 0.1943 & 0.1647 & 0.5449
& 0.0677 & 0.1849 & 0.1860 & 0.5851
& 0.0445 & 0.1830 & 0.2048 & 0.6422 \\
ELLMob
& 0.0437 & 0.0751 & 0.2159 & 0.4830
& 0.0401 & 0.0750 & 0.2341 & 0.5428
& 0.0364 & 0.0861 & 0.2875 & 0.6336 \\
\midrule
\textbf{MotiveMob}
& 0.0213 & \textbf{0.0204} & \textbf{0.0868} & \textbf{0.4498}
& 0.0256 & \textbf{0.0273} & \underline{0.1103} & \textbf{0.5137}
& 0.0273 & \textbf{0.0303} & \underline{0.1153} & \textbf{0.6060} \\
MotiveMob w/o $P_M$
& 0.0201 & 0.0311 & 0.1219 & 0.4627
& \underline{0.0200} & 0.0354 & 0.1471 & \underline{0.5153}
& 0.0225 & 0.0454 & 0.1714 & 0.6253 \\
MotiveMob w/o $\mathcal{R}_u$
& \textbf{0.0142} & 0.0385 & 0.1023 & 0.5082
& \textbf{0.0146} & 0.0357 & 0.1243 & 0.5700
& \textbf{0.0154} & 0.0445 & 0.1485 & 0.6149 \\
MotiveMob w/o $\mathcal{H}_{u,d}$
& 0.0289 & \underline{0.0258} & 0.1193 & 0.4777
& 0.0351 & \underline{0.0309} & 0.1443 & 0.5394
& 0.0306 & 0.0495 & 0.1385 & 0.6272 \\
MotiveMob w/o Speed
& 0.0633 & 0.0287 & \underline{0.0870} & \underline{0.4527}
& 0.0698 & 0.0373 & 0.1145 & 0.5230
& 0.0775 & 0.0454 & 0.1164 & 0.6163 \\
MotiveMob w/o Rep.
& \underline{0.0181} & 0.0262 & 0.0925 & 0.4535
& 0.0227 & 0.0329 & \textbf{0.1079} & 0.5216
& \underline{0.0210} & \underline{0.0326} & \textbf{0.1122} & \underline{0.6082} \\
\bottomrule
\end{tabular}%
}
\end{subtable}

\end{table*}




\subsection{Analysis of Prediction Modules}

\paragraph{Step-level Motivation Predictor.}
The \emph{w/o $P_M$} rows in Tables~\ref{tab:seen-users-temporal} and~\ref{tab:unseen-users-temporal} evaluate a direct predictor that generates the next mobility state from the current context without first inferring a step-level motivation. This variant retains a learned termination decision by predicting either the next state or \texttt{END}; the comparison therefore isolates the effect of motivation-mediated state prediction rather than removing trajectory termination.

Removing $P_M$ increases both SI and DARD in all six evaluation settings. The mean relative increases in SI and DARD are 44.9\% and 30.1\% for seen users and 44.0\% and 40.8\% for unseen users, respectively. In contrast, the effects of $P_M$ on the spatial metrics are mixed: the mean absolute relative change in STVD is only 2.6\%, while $P_M$ reduces SD divergence in two of the three seen-user settings but increases it in all three unseen-user settings. These improvements in SI and DARD are not simply explained by termination behavior or average trajectory length. As detailed in Appendix~\ref{app:termination_analysis}, the direct predictor generates 7.52 states per day on average, compared with 8.08 for MotiveMob and 6.82 in the ground truth. Although the direct predictor is closer to the ground-truth mean length, it still performs worse on SI and DARD. These results suggest that step-level motivation primarily helps preserve temporal and activity semantics by narrowing spatially plausible alternatives according to behavioral intent, which helps determine when the next move should occur and how it fits into the user's daily activity sequence. To visualize the effect of motivation predictor, Appendix~\ref{app:motivation-predictor-cases} presents representative trajectory-level case studies comparing MotiveMob w/ and w/o motivation predictor.

\paragraph{IWM for State Prediction.}
We compare the IWM-based state predictor, trained on both observed and alternative transitions, with two alternatives while keeping the motivation predictor $P_M$ fixed and inference-time rollout validation disabled. The positive-only predictor is trained solely on observed transitions, whereas the rule-based predictor retrieves observed mobility transitions compatible with each predicted motivation instead of learning a generative transition model.

\begin{table*}[t]
\centering
\caption{State predictor comparison without rollout validation for seen and unseen users across the three temporal settings. All values are JSD; lower is better. The best result in each column of each panel is shown in bold, and the runner-up is underlined.}
\label{tab:state-predictor-comparison}
\setlength{\tabcolsep}{4pt}
\renewcommand{\arraystretch}{1.12}

\resizebox{\textwidth}{!}{%
\begin{tabular}{l*{12}{c}}
\toprule
\multirow{2}{*}{State predictor $P_S$}
& \multicolumn{4}{c}{Seen period}
& \multicolumn{4}{c}{Regular unseen period}
& \multicolumn{4}{c}{COVID-19 unseen period} \\
\cmidrule(lr){2-5}
\cmidrule(lr){6-9}
\cmidrule(l){10-13}
& SD & SI & DARD & STVD
& SD & SI & DARD & STVD
& SD & SI & DARD & STVD \\
\midrule

\multicolumn{13}{l}{\textbf{Seen users}} \\
\addlinespace[2pt]

\textbf{MotiveMob} (IWM $P_S$)
& \textbf{0.0241} & 0.0159 & \underline{0.0857} & \textbf{0.3841}
& \textbf{0.0307} & 0.0212 & \underline{0.1054} & \underline{0.4558}
& \underline{0.0880} & \underline{0.0309} & 0.1632 & \underline{0.5745} \\

\quad w/ positive-only $P_S$
& \underline{0.0328} & \underline{0.0115} & \textbf{0.0707} & \underline{0.3870}
& \underline{0.0376} & \underline{0.0185} & \textbf{0.0943} & \textbf{0.4469}
& 0.0914 & 0.0390 & \underline{0.1586} & \textbf{0.5602} \\

\quad w/ rule-based $P_S$
& 0.0374 & \textbf{0.0101} & 0.0992 & 0.4286
& 0.0405 & \textbf{0.0165} & 0.1111 & 0.4932
& \textbf{0.0687} & \textbf{0.0198} & \textbf{0.1576} & 0.6230 \\

\midrule

\multicolumn{13}{l}{\textbf{Unseen users}} \\
\addlinespace[2pt]

\textbf{MotiveMob} (IWM $P_S$)
& \underline{0.0460} & 0.0356 & \underline{0.0906} & \textbf{0.4559}
& \underline{0.0473} & 0.0422 & \underline{0.1120} & \textbf{0.5240}
& \underline{0.0597} & 0.0359 & \underline{0.1209} & \textbf{0.6109} \\

\quad w/ positive-only $P_S$
& 0.0713 & \textbf{0.0244} & \textbf{0.0878} & \underline{0.4875}
& 0.0876 & \textbf{0.0300} & 0.1181 & \underline{0.5417}
& 0.1045 & \textbf{0.0220} & \textbf{0.1165} & \underline{0.6440} \\

\quad w/ rule-based $P_S$
& \textbf{0.0388} & \underline{0.0333} & 0.0916 & 0.5663
& \textbf{0.0375} & \underline{0.0341} & \textbf{0.1097} & 0.5926
& \textbf{0.0434} & \underline{0.0317} & 0.1399 & 0.6478 \\

\bottomrule
\end{tabular}%
}
\vspace{2pt}
\end{table*}

As shown in Table~\ref{tab:state-predictor-comparison}, the IWM-based predictor achieves lower SD than the positive-only predictor in all six settings and lower STVD in four, including all three unseen-user settings, whereas the positive-only predictor achieves lower SI and DARD in five settings. Relative to the rule-based predictor, the IWM-based predictor achieves lower STVD in all six settings and lower DARD in four, whereas the rule-based predictor achieves lower SD in four and lower SI in all six. These results highlight a trade-off between matching individual mobility marginals and preserving user-conditioned spatiotemporal structure. The clearest advantage of IWM training lies in modeling the joint time--location distribution, particularly for unseen users, suggesting that alternative transitions broaden the motivation-conditioned state distribution and improve spatiotemporal generalization beyond observed-transition learning or rule-based retrieval alone. Appendix~\ref{app:motivation-predictor-sensitivity} further shows that IWM initialization provides no consistent final improvement for the motivation predictor.

\subsection{Ablation Studies}

\paragraph{User-specific information.}
Removing the behavioral report $\mathcal{R}_u$ worsens SI and STVD in all six settings, confirming its role as a long-term prior over when and where a user typically performs an activity. Its effect on DARD is weaker for seen users but more pronounced for unseen users. The mixed SD results further suggest that displacement patterns encoded in the report are less reliably calibrated under user-level shifts. Recent mobility history provides more immediate conditioning: removing $\mathcal{H}_{u,d}$ worsens SI in all settings and substantially increases DARD and SD for both user groups. Its effect on STVD is smaller than that of $\mathcal{R}_u$, indicating that recent history mainly locates the user within the current routine and constrains the next activity, its timing, and the associated displacement.

\paragraph{Inference-time rollout validation.}
Removing speed validation increases SD and STVD in all six settings, whereas its effects on SI and DARD are smaller and less uniform. Speed validation therefore primarily prevents physically implausible spatial transitions. Repetition-aware termination serves a different function: disabling it consistently worsens SI and STVD but produces smaller and more mixed changes in SD and DARD. Its primary role is therefore to prevent repeated location cycles from extending trajectories to the maximum length, rather than driving MotiveMob's overall distributional gains. Detailed resampling, correction, and termination statistics are reported in Appendix~\ref{app:inference-time-validation-termination}.

\subsection{Spatial Distribution Alignment under User and Temporal Shifts}
\begin{figure*}[t]
    \centering
    \includegraphics[width=0.9\textwidth]{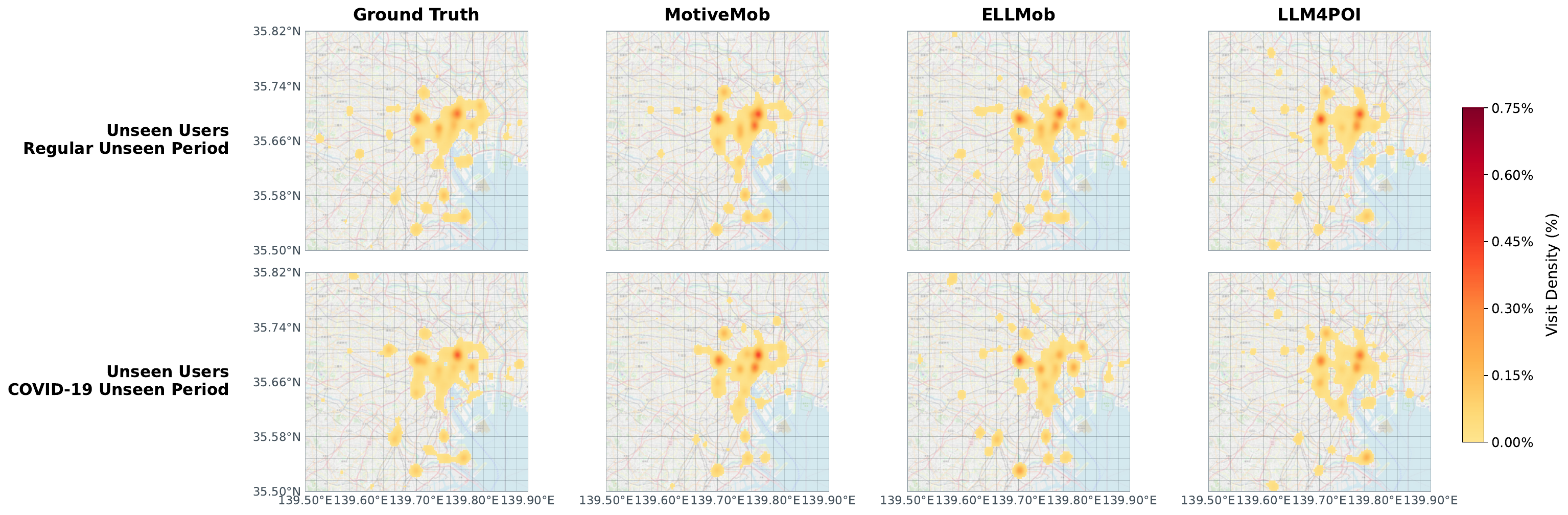}
    \caption{Ground-truth and predicted spatial mobility distributions for unseen users across two unseen periods in Tokyo. Darker red indicates higher visit density.}
    \label{fig:spatial}
\end{figure*}

As shown in Figure~\ref{fig:spatial}, all three methods recover the primary activity centers in the ground-truth distribution. Compared with LLM4POI, MotiveMob forms more compact hotspots and exhibits less peripheral dispersion, thereby better preserving the concentrated structure of the ground-truth distribution. Compared with ELLMob, MotiveMob more closely matches the ground-truth allocation of visit mass across activity regions, particularly under the COVID-19 shift. These results suggest that MotiveMob combines compact hotspot structure with more robust spatial alignment under joint user and temporal shifts. Additional spatial-distribution results are provided in Appendix~\ref{app:additional-spatial-distributions}.

\section{Conclusions and Limitations}



By combining step-level prediction of motivation proxies, motivation-conditioned state generation, and inference-time rollout validation, our motivation-first framework, \textbf{MotiveMob}, achieved the strongest overall distributional fidelity across the evaluated user groups and temporal periods. Spatial analysis further showed that MotiveMob better preserves core activity regions for unseen users under temporal shifts. Additional analyses indicate that step-level motivation prediction primarily improves temporal and activity fidelity, while IWM training with alternative transitions improves joint spatiotemporal generalization, particularly for unseen users. Together, these findings support motivation proxies as useful intermediate representations for mobility trajectory generation.

As for limitations, MotiveMob relies on a predefined motivation taxonomy and pseudo-labeled alternative transitions whose correspondence to unobserved user intentions cannot be directly verified. The current IWM augmentation also exhibits a metric-specific trade-off: improvements in the
joint time--location distribution do not consistently coincide with improvements in the step-distance distribution, suggesting that the generated alternative transitions may not simultaneously preserve local movement geometry and global spatiotemporal structure. Moreover, the step-level training objectives do not explicitly optimize trajectory-level coherence or capture the delayed effects of earlier decisions. Future work could explore adaptive motivation representations, improved augmentation strategies that better preserve both local and global mobility patterns, and hierarchical sequential objectives with trajectory-level feedback or reinforcement learning.

\subsubsection*{Acknowledgments}
This work was financially supported by JST CREST JPMJCR22M2, JSPS KAKENHI JP23K17456, JP23K28096, JP24K02996, JP25H01117, JP25K21207, JP26K03246.

\bibliography{iclr2027_conference}
\bibliographystyle{iclr2027_conference}

\clearpage
\appendix

\section{User-Specific Context Representation}
\label{app:user-context}

MotiveMob conditions both predictors on two sources of user-specific information: a long-term behavioral report $\mathcal{R}_u$ and a recent mobility history $\mathcal{H}_i$. This section specifies their textual representations.

\subsection{Behavioral Report}
\label{app:behavioral-report}

The behavioral report summarizes recurring mobility regularities derived from the user's training trajectories. It contains the typical starting
and ending check-ins on weekdays and weekends, together with frequently visited POIs and their representative visit times. The report supplied to the predictors follows the template below.

\begin{lstlisting}[style=prompt]
During weekday, you usually begin your daily trip at
<weekday-start-POI> at <HH:MM:SS> and end your daily trip at
<weekday-end-POI> at <HH:MM:SS>.

During weekend, you usually begin your daily trip at
<weekend-start-POI> at <HH:MM:SS> and end your daily trip at
<weekend-end-POI> at <HH:MM:SS>.

You usually visit:
<frequent-POI-1> at <HH:MM:SS>
<frequent-POI-2> at <HH:MM:SS>
...
\end{lstlisting}

Each POI is represented as \texttt{<category>\#<ID>}, where the ID distinguishes venues belonging to the same category.

\subsection{Recent Mobility History}
\label{app:mobility-history}

The mobility history contains up to the 30 most recent check-ins, retained in chronological order, from the three active days immediately preceding the target day:

\begin{lstlisting}[style=prompt]
<POI-1>#<ID> at <HH:MM:SS>
<POI-2>#<ID> at <HH:MM:SS>
...
<POI-K>#<ID> at <HH:MM:SS>
\end{lstlisting}

where $K\leq 30$.
\FloatBarrier

\section{Mobility Motivation Taxonomy}
\label{app:motivation-taxonomy}

Table~\ref{tab:motivation-taxonomy} summarizes the motivation taxonomy used in our experiments. Each behavioral motivation is deterministically assigned according to the category of the destination venue. The \texttt{END} label is added separately after the final check-in of each trajectory.

\begin{table*}[!htbp]
\centering
\small
\caption{Mobility motivation taxonomy and its mapping from destination
venue categories. The listed venues summarize the category groups
associated with each motivation.}
\label{tab:motivation-taxonomy}
\begin{tabularx}{\textwidth}{l X X}
\toprule
\textbf{Motivation} &
\textbf{Behavioral interpretation} &
\textbf{Destination venue categories} \\
\midrule
Return
& Returning to a residence or temporary accommodation
& Residential venues, hotels, hostels, inns, resorts, and other lodging venues \\

Personal Service
& Obtaining a personal, household, financial, repair, or travel-related service
& Salons, banks, repair shops, rental services, post offices, travel agencies, and related service venues \\

Entertainment
& Participating in leisure, cultural, or recreational activities
& Arts and entertainment, landmarks and outdoors, sports and recreation, cruises, auditoriums, and related venues \\

Dining
& Eating or drinking
& Dining and drinking venues, food courts, and cafeterias \\

Transit
& Traveling or transferring between places
& Airports, stations, public transport facilities, roads, and other transportation venues \\

Religious
& Participating in religious or spiritual activities
& Shrines, temples, churches, mosques, monasteries, and other religious venues \\

Education
& Attending educational or learning activities
& Schools, universities, libraries, classrooms, laboratories, and training institutions \\

Work
& Engaging in occupational or professional activities
& Offices, government facilities, administrative buildings, and other professional workplaces \\

Social
& Participating in social, civic, or community activities
& Event venues, wedding halls, funeral homes, and community or civic venues \\

Shopping
& Purchasing goods
& Retail venues and food-and-beverage retailers \\

Healthcare
& Receiving medical or health-related care
& Hospitals, clinics, pharmacies, and other health and medicine venues \\

\texttt{END}
& Terminating the daily trajectory
& Added after the final check-in; not associated with a destination venue \\
\bottomrule
\end{tabularx}
\end{table*}
\FloatBarrier

\section{Calibration of the Speed-Feasibility Threshold}
\label{app:speed-threshold}

We calibrate the speed threshold using only the observed transitions of the 25 seen users from January 1 to October 31, 2019. After removing one \texttt{START} row from each of the 3,559 training trajectories, the data contain 27,504 consecutive check-in transitions. For a transition from $(l_i,t_i)$ to $(l_{i+1},t_{i+1})$, its implied speed is computed as
\begin{equation}
v_i =
\frac{d_{\mathrm{hav}}(l_i,l_{i+1})}
{\max(\Delta q_i,1)\times 10/60},
\end{equation}
where $d_{\mathrm{hav}}$ is the Haversine distance in kilometers and $\Delta q_i$ is the difference between the corresponding 10-minute time slots. This definition is identical to that used by the speed-feasibility check during inference.

We exclude 14 transitions involving non-spatial states, such as \texttt{Home (private)}, and 418 transitions with displacements exceeding 100 km, leaving 27,072 transitions for calibration. The 100 km criterion prevents a small number of extreme long-distance transitions from dominating the threshold estimate and is applied only during calibration, not as a distance constraint during inference.

Because the resulting speed distribution remains strongly right-skewed, we apply the transformation
\begin{equation}
y_i =
\log\left(1+\frac{v_i}{v_0}\right),
\qquad
v_0=1\ \mathrm{km/h},
\end{equation}
which retains zero-speed transitions while compressing the upper tail.

\begin{figure*}[!htbp]
    \centering
    \includegraphics[width=0.92\textwidth]
    {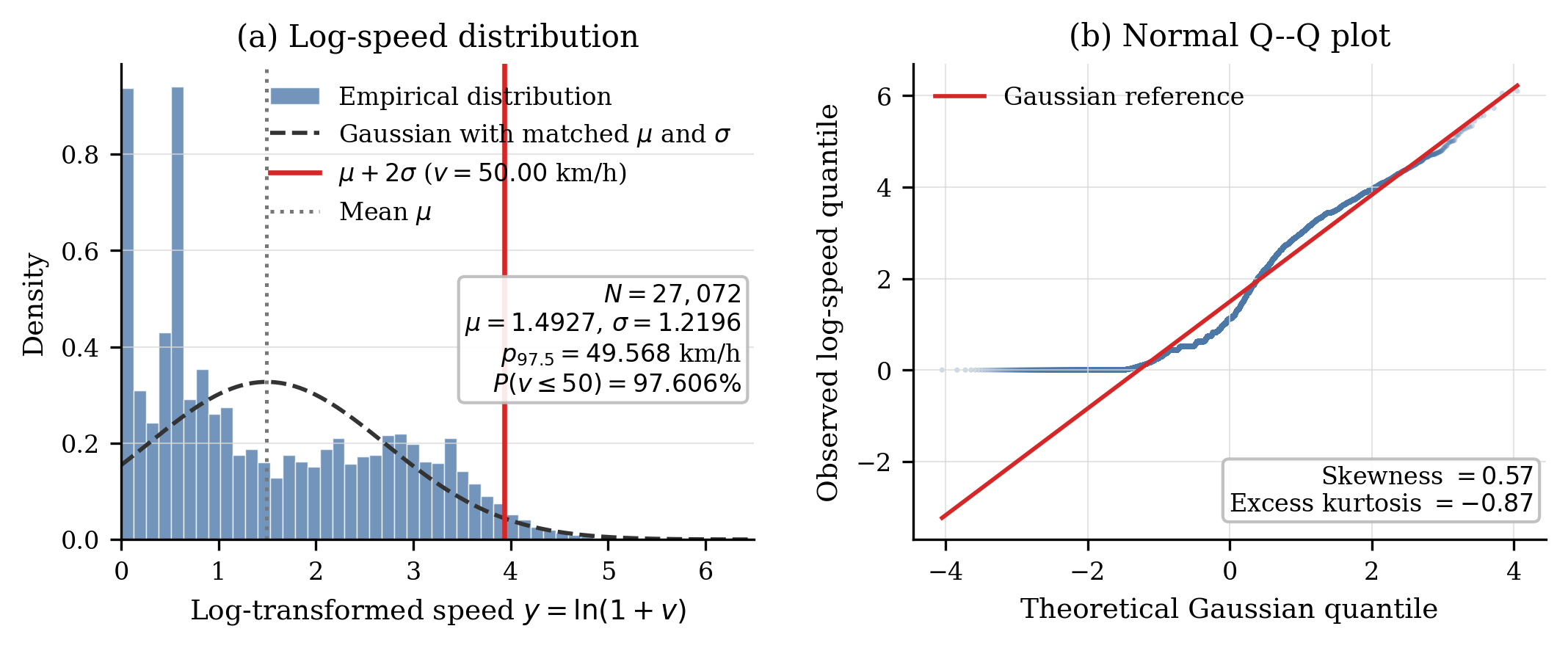}
    \caption{
    Distributional diagnostics for the log-transformed transition speeds used to calibrate $v_{\max}$.
    (a) Empirical distribution and a Gaussian distribution with the same mean and standard deviation. The red line marks the two-standard-deviation upper bound.
    (b) Normal Q--Q plot showing the remaining deviations from Gaussianity, particularly in the tails.
    }
    \label{fig:log1p-speed-distribution}
\end{figure*}

As shown in Figure~\ref{fig:log1p-speed-distribution}, the transformation substantially reduces the skewness of the speed distribution, although the Q--Q plot still shows deviations from exact Gaussianity.
\begin{equation}
\mu_y=1.4927,
\qquad
\sigma_y=1.2196,
\end{equation}
giving the following upper bound on the original speed scale:
\begin{equation}
v_0\left[\exp(\mu_y+2\sigma_y)-1\right]
=50.003\ \mathrm{km/h}.
\end{equation}

This value closely matches the empirical 97.5th percentile,
\begin{equation}
Q_{0.975}\!\left(\{v_i\}\right)
=49.568\ \mathrm{km/h}.
\end{equation}
We therefore set $v_{\max}=50$ km/h, which covers 97.606\% of the cleaned training transitions. The empirical percentile provides the primary calibration criterion, while the log-space two-standard-deviation bound serves as a consistent distributional check. The resulting threshold is fixed across all user groups and temporal periods without using any unseen-user or unseen-period data.
\FloatBarrier

\section{Implementation Details and Configurations}
\label{app:training-inference}

We use Llama-3.1-8B-Instruct as the base model for the LLM judge, the state predictor $P_S$, and the motivation predictor $P_M$. The LLM judge remains frozen during alternative transition construction, whereas $P_S$ and $P_M$ are independently adapted from the base model using QLoRA. This appendix provides implementation details and configurations for three stages: alternative transition construction, predictor training, and autoregressive trajectory inference.

\subsection{Alternative Transition Construction}
\label{app:alternative-construction}

Table~\ref{tab:alternative-construction-config} summarizes the data, candidate-generator, and LLM-judge configurations used to construct alternative transitions. For each observed transition, we sample three distinct motivations other than the observed one. The candidate generator first proposes possible next states for each alternative motivation, after which the LLM judge selects the most plausible candidate under the user's behavioral context.

\begin{table*}[!htbp]
\centering
\caption{Configuration for alternative-transition construction.}
\label{tab:alternative-construction-config}
\small
\setlength{\tabcolsep}{6pt}
\renewcommand{\arraystretch}{1.10}

\begin{tabular}{@{}L{0.27\textwidth}L{0.67\textwidth}@{}}
\toprule
Parameter & Setting \\
\midrule

Construction data
& 25 seen users from January--October 2019, containing 3,559 daily trajectories and 31,063 observed transition instances \\
\addlinespace[2pt]

Alternative motivations
& Three distinct motivations per observed transition, sampled without replacement while excluding the observed motivation \\
\addlinespace[2pt]

Prepared alternative cases
& 93,189 cases before candidate generation and judging; 31,063 judged alternative transitions are used for balanced state-predictor training \\
\addlinespace[2pt]

Candidate-generator architecture
& Three-layer pre-layer-normalized Transformer encoder with hidden dimension 128, 4 attention heads, feed-forward dimension 256, and dropout 0.1 \\
\addlinespace[2pt]

Prediction heads
& Separate classification heads for location subcategory, spatial grid, and arrival-time bin \\
\addlinespace[2pt]

Candidate-generator objective
& Equal-weighted cross-entropy losses for the subcategory, spatial-grid, and time-bin prediction heads \\
\addlinespace[2pt]

Candidate-generator split
& 90\% training and 10\% validation dates within each user, split by complete user--date groups with seed 42 \\
\addlinespace[2pt]

Candidate-generator optimization
& AdamW with learning rate $1\times10^{-3}$, weight decay $1\times10^{-4}$, batch size 2,048, and 30 training epochs \\
\addlinespace[2pt]

Learning-rate schedule
& Two warm-up epochs followed by cosine decay with a minimum learning-rate ratio of 0.05 \\
\addlinespace[2pt]

Additional regularization
& Gradient clipping at 1.0 and label smoothing of 0.05 \\
\addlinespace[2pt]

Checkpoint selection
& The checkpoint with the highest mean top-1 validation accuracy over the subcategory, spatial-grid, and time-bin heads \\
\addlinespace[2pt]

Candidate enumeration
& Top 8 subcategories, top 8 spatial grids, and top 8 time bins are combined into at most 50 preliminary triples, from which at most 10 candidates are retained for judging \\
\addlinespace[2pt]

Candidate-scoring weights
& $\alpha_{\mathrm{sub}}=0.8$, $\alpha_{\mathrm{grid}}=0.9$, $\alpha_{\mathrm{time}}=0.55$, $\alpha_{\mathrm{user}}=1.0$ \\
\addlinespace[2pt]

Candidate filtering
& Motivation-strict filtering and behavioral-report-based distance capping are disabled; equal-time alternative transitions are not allowed \\
\addlinespace[2pt]

Candidate context
& The previous three active mobility days, the preceding check-ins on the target day, the current state, the alternative motivation, and the user's long-term behavioral report \\
\addlinespace[2pt]

LLM judge
& Frozen Llama-3.1-8B-Instruct loaded with 4-bit NF4 quantization, double quantization, and bfloat16 computation when supported \\
\addlinespace[2pt]

Judge decoding
& Greedy decoding with at most 8 newly generated tokens \\
\addlinespace[2pt]

Judge decision
& The judge must select one candidate when the candidate set is nonempty. An invalid or unparsable output defaults to the highest-ranked candidate; no alternative transition is retained when the candidate set is empty \\
\addlinespace[2pt]

$\rho$
& 0.33 \\
\addlinespace[2pt]

Construction seed
& 3407 \\

\bottomrule
\end{tabular}
\end{table*}

The judge receives the target date and weekend indicator, the user's long-term behavioral report, recent mobility history, earlier check-ins on the target day, the current location and time, the alternative motivation, and the candidate states. Each candidate is represented by its location, coordinates, arrival time, activity category, distance from the current state, and candidate-generator score. The judge is instructed to return only the index of the candidate that is most consistent with temporal validity, spatial feasibility, the specified motivation, and the user's historical behavior.

\subsection{Predictor Training}
\label{app:predictor-training}

The state predictor and motivation predictor are trained independently. The state predictor is trained on a balanced mixture of observed and judged alternative transitions, preventing the larger set of initially constructed alternatives from dominating its training objective. The motivation predictor is trained only on observed step-level motivations and trajectory-termination labels. Table~\ref{tab:predictor-training-config} reports their training configurations.

\begin{table*}[!htbp]
\centering
\caption{Training configurations for the state predictor $P_S$ and motivation predictor $P_M$.}
\label{tab:predictor-training-config}
\small
\setlength{\tabcolsep}{5pt}
\renewcommand{\arraystretch}{1.10}

\begin{tabular}{@{}L{0.20\textwidth}L{0.37\textwidth}L{0.37\textwidth}@{}}
\toprule
Parameter & State predictor $P_S$ & Motivation predictor $P_M$ \\
\midrule

Initialization
& Llama-3.1-8B-Instruct
& Independently initialized from Llama-3.1-8B-Instruct \\
\addlinespace[2pt]

Training data
& 31,063 observed transitions and 31,063 judged alternative transitions
& 34,622 observed instances, comprising 31,063 step-level motivations and 3,559 \texttt{END} labels \\
\addlinespace[2pt]

Observed-to-alternative ratio
& $1{:}1$
& Not applicable \\
\addlinespace[2pt]

Training objective
& Assistant-token-only causal language-modeling loss
& Assistant-token-only causal language-modeling loss \\
\addlinespace[2pt]

Quantization
& 4-bit NF4 with double quantization and bfloat16 computation
& 4-bit NF4 with double quantization and bfloat16 computation \\
\addlinespace[2pt]

QLoRA configuration
& Rank $r=16$, scaling factor $\alpha=16$, dropout 0, and no bias
& Rank $r=16$, scaling factor $\alpha=16$, dropout 0, and no bias \\
\addlinespace[2pt]

LoRA target modules
& \texttt{q\_proj}, \texttt{k\_proj}, \texttt{v\_proj}, \texttt{o\_proj}, \texttt{gate\_proj}, \texttt{up\_proj}, and \texttt{down\_proj}
& \texttt{q\_proj}, \texttt{k\_proj}, \texttt{v\_proj}, \texttt{o\_proj}, \texttt{gate\_proj}, \texttt{up\_proj}, and \texttt{down\_proj} \\
\addlinespace[2pt]

Optimizer
& 8-bit paged AdamW
& 8-bit paged AdamW \\
\addlinespace[2pt]

Learning rate
& $1\times10^{-4}$
& $2\times10^{-4}$ \\
\addlinespace[2pt]

Weight decay
& 0.01
& 0.01 \\
\addlinespace[2pt]

Learning-rate schedule
& Linear decay with 200 warm-up steps
& Linear decay with 200 warm-up steps \\
\addlinespace[2pt]

Batch size
& 2 per GPU with 8 gradient-accumulation steps on 8 GPUs, giving an effective batch size of 128
& 2 per GPU with 8 gradient-accumulation steps on 8 GPUs, giving an effective batch size of 128 \\
\addlinespace[2pt]

Training budget
& 4,000 optimization steps
& 12 epochs; the checkpoint at step 3,084 is used for inference \\
\addlinespace[2pt]

Maximum sequence length
& 2,048 tokens
& 2,048 tokens \\
\addlinespace[2pt]

Mobility-history context
& Up to 30 check-ins from the three active days preceding the target day
& Up to 30 check-ins from the three active days preceding the target day \\
\addlinespace[2pt]

Training seed
& 3407
& 3407 \\

\bottomrule
\end{tabular}
\end{table*}

\subsection{Trajectory Inference}
\label{app:trajectory-inference}

Trajectory generation starts from a \texttt{START} state and alternates between motivation prediction and state prediction. The motivation predictor either produces the next behavioral motivation or returns \texttt{END}. Given a non-terminal motivation, the state predictor generates a candidate location and arrival time. The candidate is then validated before being appended to the trajectory. Table~\ref{tab:trajectory-inference-config} summarizes the inference configuration.

\begin{table*}[t]
\centering
\caption{Configuration for autoregressive trajectory inference.}
\label{tab:trajectory-inference-config}
\small
\setlength{\tabcolsep}{6pt}
\renewcommand{\arraystretch}{1.10}

\begin{tabular}{@{}L{0.27\textwidth}L{0.67\textwidth}@{}}
\toprule
Parameter & Setting \\
\midrule

Initial state
& A \texttt{START} state; the first motivation and check-in are generated without applying a travel-speed constraint because no previous spatial state is available \\
\addlinespace[2pt]

Mobility-history context
& Up to 30 check-ins from the three active days preceding the target day \\
\addlinespace[2pt]

Motivation decoding
& Greedy decoding with at most 8 newly generated tokens \\
\addlinespace[2pt]

State decoding
& Sampling with temperature 0.2, top-$p$ 0.9, and at most 32 newly generated tokens \\
\addlinespace[2pt]

State-generation attempts
& At most 8 proposals per step, consisting of one initial proposal and up to seven retries \\
\addlinespace[2pt]

Retry decoding
& Temperature increases from 0.3 to 0.9 in increments of 0.1; top-$p$ remains 0.9 and the maximum output length remains 32 tokens \\
\addlinespace[2pt]

Output validation
& The generated output must be parseable as a valid location--time state with a recognized location identifier and timestamp \\
\addlinespace[2pt]

Temporal validation
& The generated arrival time must not precede the current time; generation terminates when the predicted state crosses into the next mobility day \\
\addlinespace[2pt]

Travel-speed calculation
& Haversine distance divided by the elapsed time between consecutive states, with time represented using 10-minute slots \\
\addlinespace[2pt]

Speed threshold
& $v_{\max}=50~\mathrm{km/h}$; its selection is described in Appendix~\ref{app:speed-threshold} \\
\addlinespace[2pt]

Speed-based retry
& A proposal exceeding $v_{\max}$ is rejected and regenerated using the retry schedule above \\
\addlinespace[2pt]

Re-grounding after retries
& If all eight proposals violate the speed constraint, the final proposal is re-grounded, when possible, to a location of the same broad activity category that appears in the user's history or behavioral report and satisfies $v_{\max}$; feasible locations are ranked by time-of-day similarity with historical frequency used as a tie-breaker \\
\addlinespace[2pt]

Re-grounding failure
& If no feasible location is available for re-grounding, the final state-predictor proposal is retained \\
\addlinespace[2pt]

Additional fallbacks
& The legacy distance-based fallback and history-trajectory replacement fallback are disabled \\
\addlinespace[2pt]

Repetition termination
& Activated after 15 accepted check-ins. Generation terminates before appending a candidate if the candidate-augmented location sequence contains a block of length 1--4 repeated for at least three consecutive rounds \\
\addlinespace[2pt]

Online repetition guard
& Disabled; repetition is handled only by the termination rule above \\
\addlinespace[2pt]

Maximum trajectory length
& 25 generated check-ins \\

\bottomrule
\end{tabular}
\end{table*}
\FloatBarrier

\section{Prompt Templates}
\label{app:prompts}

All prompts are rendered using the chat template of Llama-3.1-8B-Instruct. Text enclosed by angle brackets below denotes sample-specific content.

\subsection{LLM Judge for Alternative Transition Construction}
\label{app:judge-prompt}

For each alternative motivation, the lightweight Transformer produces a candidate set of next states. The LLM judge selects one candidate without generating a new location or timestamp.

\noindent\textbf{System prompt}\par
\vspace{1pt}
\begin{lstlisting}[style=prompt]
You are a careful urban mobility candidate judge.
You must choose exactly one candidate next check-in from a provided candidate list.
Do not invent a new place or time.
\end{lstlisting}

\noindent\textbf{User prompt}\par
\vspace{1pt}
\begin{lstlisting}[style=prompt]
<INPUT 0> Behavior Pattern:
<BEHAVIORAL_REPORT>

<INPUT 1> Recent History Before Today:
<RECENT_HISTORY_OR_NONE>

<INPUT 2> Earlier Check-ins Today:
<EARLIER_CHECKINS_OR_NONE>

<INPUT 3> Current Location & Time:
<CURRENT_STATE> [lat=<LATITUDE>, lng=<LONGITUDE>]

<INPUT 4> Target Motivation for the NEXT transition:
<ALTERNATIVE_MOTIVATION>

<INPUT 5> Date and Weekend Flag:
date=<YYYY-MM-DD>, is_weekend=<0_OR_1>

<CANDIDATES>
1. <LOCATION>#<ID> at <HH:MM:SS>
   [transformer_score=<SCORE>;
    top_category=<CATEGORY>;
    candidate_motivation=<MOTIVATION>;
    lat=<LATITUDE>, lng=<LONGITUDE>;
    distance_from_current_km=<DISTANCE>]
...
K. <LOCATION>#<ID> at <HH:MM:SS> [<METADATA>]

Task:
Choose the single best candidate for the next check-in. Prefer candidates
that fit the target motivation, the user's recent behavior, the current
state, and temporal order.

Output format:
Output ONLY the candidate number, such as 1 or 7. Do not output any other text.
\end{lstlisting}

\subsection{State Predictor}
\label{app:state-prompt}

The same state-prediction prompt is used for observed and alternative transitions during training. At inference time, the target motivation is the output of $P_M$.

\noindent\textbf{System prompt}\par
\vspace{1pt}
\begin{lstlisting}[style=prompt]
You are an urban mobility transition model.
Given a behavior pattern, recent history checkins, current location and
time, current motivation, today's date, and is_weekend, predict the next
location and time.
Follow the required output format exactly.
\end{lstlisting}

\noindent\textbf{User prompt}\par
\vspace{1pt}
\begin{lstlisting}[style=prompt]
<INPUT 0> (Behavior Pattern - long-term prior):
<BEHAVIORAL_REPORT>

<INPUT 1> (Recent History Checkins):
<RECENT_HISTORY>

<INPUT 2> (Current Location & Time):
<CURRENT_STATE>

<INPUT 3> (Target Motivation for the NEXT transition):
<MOTIVATION>

<INPUT 4> (Today's Date):
<YYYY-MM-DD>

<INPUT 5> (Is Weekend, 1=yes 0=no):
<0_OR_1>

Task:
Predict the NEXT location and time.

Rules:
- Recent history and current state are more important than the behavior pattern.
- Use the behavior pattern only as a general long-term reference.
- The next timestamp must not be earlier than the current timestamp under
  the day definition starting at 04:00:00.

Output format (IMPORTANT):
<Location>#<ID> at HH:MM:SS
Do NOT output any extra text.

Example:
Park#12 at 09:10:00
\end{lstlisting}

\subsection{Motivation Predictor}
\label{app:motivation-prompt}

The motivation predictor does not observe the next state or any subsequent state. It predicts either the motivation underlying the next transition or the termination label \texttt{END}.

\noindent\textbf{System prompt}\par
\vspace{1pt}
\begin{lstlisting}[style=prompt]
You are an urban mobility intention inference model.
Given a behavior pattern, recent history Checkins, the current location
and time, today's date, and is_weekend, infer the motivation for the next
transition from the current state.
Output exactly one label from the provided list.
\end{lstlisting}

\noindent\textbf{User prompt}\par
\vspace{1pt}
\begin{lstlisting}[style=prompt]
<INPUT 0> (Behavior Pattern - long-term prior):
<BEHAVIORAL_REPORT>

<INPUT 1> (Recent History Checkins):
<RECENT_HISTORY>

<INPUT 2> (Current Location & Time):
<CURRENT_STATE>

<INPUT 3> (Today's Date):
<YYYY-MM-DD>

<INPUT 4> (Is Weekend, 1=yes 0=no):
<0_OR_1>

Task:
Infer the motivation for the next transition.

Rules:
- Recent history is more important than the behavior pattern.

Allowed labels (choose exactly ONE):
- Return
- Personal Service
- Entertainment
- Dining
- Transit
- Religious
- Education
- Work
- Social
- Shopping
- Healthcare
- END

Output format (IMPORTANT):
- Output EXACTLY ONE line.
- The line MUST be exactly one of the allowed labels.
- Do NOT output any extra text.
\end{lstlisting}
\FloatBarrier

\section{Motivation-Guided Trajectory Generation Pseudo Codes}
\label{app:trajectory-generation}
This section presents the pseudocode for the complete inference-time rollout, including feasibility-aware resampling, repetition-based termination, and mobility-grounded recovery.
\begin{algorithm}[!htbp]
\caption{Motivation-guided trajectory generation}
\label{alg:motivation-guided-rollout}
\begin{algorithmic}[1]
\Require Models $P_M$ and $P_S$; user context $x_u$; observed locations $\mathcal{O}_u$
\Require $v_{\max}=50$ km/h; $R=8$; $L_{\min}=15$; $L_{\max}=25$
\Ensure Generated trajectory $\mathcal{T}$

\State $\mathcal{T}\gets[\,]$
\State $c_0\gets\Call{BuildContext}{x_u,\mathtt{START},\mathcal{T}}$

\For{$i=0,\ldots,L_{\max}-1$}
    \State $\hat{m}_i\gets
    \displaystyle\arg\max_{y\in\mathcal{M}\cup\{\mathtt{END}\}}
    p_{\theta_M}(y\mid c_i)$

    \If{$\hat{m}_i=\mathtt{END}$}
        \State \Return $\mathcal{T}$
    \EndIf

    \State $s^\star\gets\varnothing$
    \State $\tilde{s}^{\mathrm{last}}\gets\varnothing$

    \For{$r=1,\ldots,R$}
        \If{$r=1$}
            \State $T_r\gets0.2$
        \Else
            \State $T_r\gets0.3+0.6(r-2)/(R-2)$
        \EndIf

        \State $\tilde{s}\sim
        p_{\theta_S}^{(T_r,\;p=0.9)}
        (\cdot\mid c_i,\hat{m}_i)$
        \State $\tilde{s}\gets\Call{ParseState}{\tilde{s}}$

        \If{$\tilde{s}=\varnothing$}
            \State \textbf{continue}
        \EndIf

        \State $\tilde{s}^{\mathrm{last}}\gets\tilde{s}$

        \If{$i>0$ and
        $\Call{ArrivalTime}{\tilde{s}}<
        \Call{ArrivalTime}{s_{i-1}}$}
            \State \textbf{continue}
        \EndIf

        \If{$i>0$ and
        $\Call{ImpliedSpeed}{s_{i-1},\tilde{s}}>v_{\max}$}
            \State \textbf{continue}
        \EndIf

        \State $s^\star\gets\tilde{s}$
        \State \textbf{break}
    \EndFor

    \If{$s^\star=\varnothing$}
        \If{$\tilde{s}^{\mathrm{last}}=\varnothing$}
            \State \Return $\mathcal{T}$
        \EndIf

        \State $s^\star\gets
        \Call{ReGround}{
            \tilde{s}^{\mathrm{last}},
            s_{i-1},
            \mathcal{O}_u,
            v_{\max}}$

        \If{$s^\star=\varnothing$}
            \State \Return $\mathcal{T}$
        \EndIf
    \EndIf

    \State $\mathcal{L}^{+}\gets
    \Call{Locations}{\mathcal{T}\mathbin{\|}[s^\star]}$

    \If{$|\mathcal{T}|\ge L_{\min}$ and
    $\Call{RepeatedCycle}{\mathcal{L}^{+},4,3,\{1,2,3,4\}}$}
        \State \Return $\mathcal{T}$
    \EndIf

    \State $\mathcal{T}\gets\mathcal{T}\mathbin{\|}[s^\star]$
    \State $s_i\gets s^\star$
    \State $c_{i+1}\gets\Call{BuildContext}{x_u,s_i,\mathcal{T}}$
\EndFor

\State \Return $\mathcal{T}$
\end{algorithmic}
\end{algorithm}

\begin{algorithm}[!htbp]
\caption{Repetition-aware termination}
\label{alg:repetition-stop}
\begin{algorithmic}[1]
\Function{RepeatedCycle}{$\mathcal{L},B,K,\mathcal{P}$}
    \State $n\gets|\mathcal{L}|$

    \For{$b=1,\ldots,\min\{B,\lfloor n/K\rfloor\}$}
        \For{$a=1,\ldots,n-Kb+1$}
            \State $W\gets\mathcal{L}[a:a+Kb-1]$
            \State $Q\gets W[1:b]$

            \If{$W=Q^K$ and
            $\Call{PrimitivePeriod}{W}\in\mathcal{P}$}
                \State \Return \textbf{true}
            \EndIf
        \EndFor
    \EndFor

    \State \Return \textbf{false}
\EndFunction
\end{algorithmic}
\end{algorithm}

\begin{algorithm}[!htbp]
\caption{Mobility-grounded recovery after retry exhaustion}
\label{alg:reground}
\begin{algorithmic}[1]
\Function{ReGround}{$\tilde{s},s_{\mathrm{prev}},\mathcal{O}_u,v_{\max}$}
    \State $(\tilde{\ell},\tilde{t})\gets\Call{SplitState}{\tilde{s}}$
    \State $a\gets\Call{ActivityCategory}{\tilde{\ell}}$
    \State $\mathcal{C}\gets
    \Call{ObservedCandidates}{\mathcal{O}_u,a}$
    \State $\mathcal{F}\gets\emptyset$

    \ForAll{$(\ell,t^{\mathrm{obs}})\in\mathcal{C}$}
        \State $s'\gets(\ell,\tilde{t})$

        \If{$s_{\mathrm{prev}}=\mathtt{START}$ or
        $\Call{ImpliedSpeed}{s_{\mathrm{prev}},s'}\leq v_{\max}$}
            \State $\mathcal{F}\gets
            \mathcal{F}\cup\{(\ell,t^{\mathrm{obs}})\}$
        \EndIf
    \EndFor

    \If{$\mathcal{F}=\emptyset$}
        \State \Return $\emptyset$
    \EndIf

    \State $(\ell^\star,t^\star)\gets
    \displaystyle\arg\min_{(\ell,t^{\mathrm{obs}})\in\mathcal{F}}
    \left(
        \left|
        \Call{TimeSlot}{t^{\mathrm{obs}}}
        -
        \Call{TimeSlot}{\tilde{t}}
        \right|,
        -\Call{VisitFrequency}{u,\ell}
    \right)$

    \State \Return $(\ell^\star,\tilde{t})$
\EndFunction
\end{algorithmic}
\end{algorithm}
\FloatBarrier

\section{Evaluation-Split Construction and Leakage Prevention}
\label{app:split_protocol}

We distinguish three roles of mobility trajectories in our experiments.
\emph{Parameter-training trajectories} are used to optimize the candidate
generator, the state predictor $P_S$, and the motivation predictor $P_M$.
\emph{Support trajectories} are used only to construct user-specific context
at inference time. \emph{Query trajectories} are reserved exclusively as
evaluation targets. In particular, support trajectories from unseen users
never participate in parameter optimization.

\subsection{Seen Users in the Seen Period}

For each of the 25 seen users, we randomly divide the daily trajectories
collected between January and October 2019 into disjoint training and query
subsets using an approximately 80/20 ratio. After per-user rounding, this
produces 3,559 parameter-training trajectories and 908 held-out query
trajectories. We verify that the two subsets contain no common
$(\text{user},\text{date})$ pair.

Both $P_S$ and $P_M$ are trained exclusively from the 3,559 training
trajectories. Each trajectory contributes one \texttt{END} instance to the
training data of $P_M$; therefore, the 3,559 \texttt{END} instances reported
in Table~5 correspond exactly to the parameter-training trajectories and do
not include any of the 908 evaluation trajectories.

The candidate generator further divides the same 3,559 trajectories into
90\% training dates and 10\% validation dates for checkpoint selection. This
is an internal training/validation division and is independent of the
908-trajectory final evaluation set. We therefore treat the seen-user,
seen-period setting as an \emph{in-distribution held-out} evaluation rather
than a user- or time-shift generalization setting.

\subsection{Inference Context for Unseen Users}

The 50 unseen users are disjoint from the 25 seen users, and none of their
trajectories is used to optimize the candidate generator, $P_S$, or $P_M$.
For the unseen-user, seen-period evaluation, their January--October 2019
trajectories are divided into 5,431 support trajectories and 1,345 held-out
query trajectories. These subsets contain no common
$(\text{user},\text{date})$ pair.

For an unseen user $u$, the behavioral report $R_u$ is constructed only from
that user's support trajectories. For a query trajectory $\tau_{u,d}$, the
recent mobility history $H_{u,d}$ contains up to 30 check-ins from the three
most recent active support days strictly preceding the target date $d$. The
target trajectory $\tau_{u,d}$ and all query trajectories are excluded from
both report and history construction.

Thus, ``unseen user'' means unseen during parameter learning, rather than a
zero-history cold-start setting. The corresponding experiment evaluates
generalization to new users given a limited, inference-only mobility context.

\subsection{Context Used in the Temporal-Shift Settings}

For the regular unseen period, spanning November 2019 through February 2020,
the behavioral report is constructed from January--October 2019 support
trajectories. The report window therefore precedes the complete evaluation
period. The recent history $H_{u,d}$ is selected from a support pool disjoint
from the query trajectories and contains only observations strictly preceding
$d$.

For the COVID-19 period, spanning April through October 2020, the behavioral
report is updated using only mobility observations from March 2020. The recent
history is again restricted to support trajectories strictly preceding the
target date. Consequently, neither the target-day trajectory nor a future
trajectory enters $R_u$ or $H_{u,d}$ in either temporal-shift evaluation.

The two seen-period settings use an IID trajectory holdout rather than a
chronological forecasting split. In these settings, $R_u$ is a static profile
constructed from the user's complete January--October support subset.
Accordingly, it may summarize a support trajectory dated after a particular
query date, but it never contains the query trajectory itself or any other
trajectory from the query subset. We therefore do not interpret the
seen-period results as causal future forecasting. By contrast, $H_{u,d}$ is
always restricted to support observations preceding $d$.

\begin{table}[t]
\centering
\caption{Data separation and information available under the evaluation protocol.}
\label{tab:split_protocol}
\small
\setlength{\tabcolsep}{4pt}
\renewcommand{\arraystretch}{1.12}
\begin{tabularx}{\linewidth}{@{}p{0.27\linewidth}X@{}}
\toprule
Protocol component & Data usage and restriction \\
\midrule

Predictor training
&
Trajectories from 25 seen users between January and October 2019. 
For the seen-user/seen-period evaluation, 20\% of user--date
trajectories are held out and excluded from predictor training.
\\

User-level shift
&
The 50 unseen users are disjoint from the 25 users used to train
$P_M$ and $P_S$.
\\

Temporal shift
&
The regular unseen period (November 2019--February 2020) and the
COVID-19 period (April--October 2020) do not overlap with the
predictor-training period.
\\

Recent history $\mathcal{H}_{u,d}$
&
Contains at most 30 check-ins from the three active days strictly
preceding target date $d$. Target-day and future observations are
never included.
\\

Behavioral report $R_u$
&
In the temporal-shift settings, the report is constructed only from
observations preceding the evaluation period. In the seen-period settings,
it is a static aggregate profile constructed from the query-disjoint support
subset and may include support dates later than the target date.
\\

COVID-19 update
&
For targets in April--October 2020, the report is updated using
observations available through March 2020, without using any
check-ins from the target period.
\\

Re-grounding
&
Candidate locations are drawn only from the same $R_u$ and $H_{u,d}$
provided to the predictors. Re-grounding does not access the query trajectory
or any additional trajectory database.
\\

\bottomrule
\end{tabularx}
\end{table}

\subsection{Information Available to Re-grounding}

The inference-time re-grounding procedure does not access the complete
trajectory database or the ground-truth query trajectory. Its candidate pool
contains only POI--time pairs extracted from the same behavioral report $R_u$
and recent mobility history $H_{u,d}$ supplied to $P_S$ and $P_M$. Re-grounding
therefore introduces no additional user information beyond the context already
visible to the two predictors.

The two seen-period settings use non-chronological held-out splits under
the observed-period distribution. In both settings, $R_u$ is constructed
exclusively from the user's support subset, but it may aggregate behavior
from a support trajectory dated after a particular query date. Importantly,
$R_u$ contains only aggregate behavioral regularities---such as typical
starting and ending activities and frequently visited POIs with representative
visit times---rather than any complete day-level trajectory, and it never
contains the target query trajectory or any other trajectory from the query
subset. We therefore interpret the seen-user/seen-period setting as an
in-distribution held-out evaluation and the unseen-user/seen-period setting
as a user-level transfer evaluation under the observed-period distribution;
neither is intended as causal future forecasting. By contrast, $H_{u,d}$ is
always restricted to support observations strictly preceding $d$.
\FloatBarrier

\section{Osaka Results}
\label{app:osaka}
We report the evaluation results on the Osaka human check-in dataset, following the same user-level and temporal splitting protocols used for the Tokyo dataset. To evaluate user-level generalization, we use trajectories from 25 seen users to train the base model and reserve a disjoint set of 50 additional users exclusively for evaluation. No mobility records from these unseen users are available during training. To evaluate temporal generalization, check-ins collected from January to October 2019 constitute the seen training period. We further consider two unseen periods: November 2019 to February 2020, which captures a regular seasonal shift, and April to October 2020, which captures the pandemic-induced mobility shift following the onset of COVID-19 restrictions.

\begin{table*}[!htbp]
\centering
\caption{Evaluation results on seen users in Osaka across seen and unseen time periods. The upper and lower baseline sections correspond to pretraining-based and prompting-based methods, respectively. Lower values are better; the best and second-best results are highlighted in bold and underlined, respectively.}
\label{tab:osaka-seen-users-temporal-raw}
\setlength{\tabcolsep}{4pt}
\renewcommand{\arraystretch}{1.12}
\resizebox{\textwidth}{!}{%
\begin{tabular}{l*{12}{c}}
\toprule
\multirow{2}{*}{Method}
& \multicolumn{4}{c}{Seen period (1186 trajectories)}
& \multicolumn{4}{c}{Regular unseen period (466 trajectories)}
& \multicolumn{4}{c}{COVID-19 unseen period (689 trajectories)} \\
\cmidrule(lr){2-5} \cmidrule(lr){6-9} \cmidrule(l){10-13}
& SD & SI & DARD & STVD
& SD & SI & DARD & STVD
& SD & SI & DARD & STVD \\
\midrule
LSTM
& 0.0914 & 0.1375 & 0.2673 & 0.4161
& 0.1109 & 0.1698 & 0.3230 & 0.5779
& 0.1259 & 0.2073 & 0.3556 & 0.5725 \\
MHSA
& 0.0908 & 0.1357 & 0.2468 & 0.4286
& 0.1210 & 0.1817 & 0.2706 & 0.5803
& 0.1262 & 0.2152 & 0.3044 & 0.5552 \\
DeepMove
& 0.1777 & 0.1024 & 0.2348 & 0.4933
& 0.1645 & 0.1267 & 0.3180 & 0.6065
& 0.2308 & 0.1595 & 0.3340 & 0.6029 \\
GetNext
& 0.0395 & 0.2820 & 0.2149 & 0.4635
& 0.0433 & 0.2834 & 0.2057 & 0.5601
& 0.0455 & 0.2952 & 0.1953 & 0.5294 \\
TrajGAIL
& 0.2251 & 0.1256 & \underline{0.0799} & 0.5836
& 0.2458 & 0.1227 & \underline{0.1250} & 0.6576
& 0.2494 & 0.1118 & \underline{0.1137} & 0.6424 \\
DiffTraj
& 0.2002 & 0.1217 & 0.1407 & 0.6874
& 0.2092 & 0.1165 & 0.1662 & 0.6924
& 0.1869 & 0.1025 & 0.1907 & 0.6904 \\
GNPR-SID
& 0.1121 & 0.0212 & 0.0988 & 0.4642
& 0.1319 & 0.0346 & 0.1262 & 0.5786
& 0.1889 & 0.0434 & 0.1711 & 0.5956 \\
LLM4POI
& 0.0323 & \underline{0.0192} & 0.0900 & \underline{0.4093}
& 0.0463 & 0.0340 & 0.1309 & 0.5408
& 0.0424 & 0.0512 & 0.1278 & \underline{0.5139} \\
\midrule
LLMMove
& 0.0440 & 0.0453 & 0.1102 & 0.5022
& 0.0528 & 0.0518 & 0.1866 & 0.5907
& 0.0455 & 0.0630 & 0.1556 & 0.5856 \\
LLM-ZS
& 0.0352 & 0.0521 & 0.1881 & 0.5124
& 0.0300 & 0.0534 & 0.2272 & 0.5880
& 0.0274 & 0.0648 & 0.1958 & 0.5596 \\
LLM-Mob
& 0.0338 & 0.0554 & 0.2117 & 0.4977
& 0.0309 & 0.0615 & 0.2741 & 0.5901
& 0.0382 & 0.0656 & 0.2819 & 0.5777 \\
LLMob
& 0.1153 & 0.2542 & 0.1435 & 0.5276
& 0.0537 & 0.0980 & 0.1334 & 0.5827
& 0.0530 & 0.0858 & 0.1225 & 0.5433 \\
ELLMob
& \underline{0.0160} & 0.0274 & 0.0939 & 0.4127
& \underline{0.0205} & \underline{0.0281} & 0.1496 & \underline{0.5302}
& \underline{0.0209} & \underline{0.0334} & 0.1442 & 0.5405 \\
\midrule
\textbf{MotiveMob}
& \textbf{0.0038} & \textbf{0.0078} & \textbf{0.0340} & \textbf{0.4035}
& \textbf{0.0097} & \textbf{0.0165} & \textbf{0.0884} & \textbf{0.5153}
& \textbf{0.0125} & \textbf{0.0207} & \textbf{0.0723} & \textbf{0.5102} \\
\bottomrule
\end{tabular}%
}
\vspace{2pt}
\end{table*}

\begin{table*}[!htbp]
\centering
\caption{Evaluation results on unseen users in Osaka across seen and unseen time periods. The upper and lower baseline sections correspond to pretraining-based and prompting-based methods, respectively. Lower values are better; the best and second-best results are highlighted in bold and underlined, respectively.}
\label{tab:osaka-unseen-users-fixed50-temporal-raw}
\setlength{\tabcolsep}{4pt}
\renewcommand{\arraystretch}{1.12}
\resizebox{\textwidth}{!}{%
\begin{tabular}{l*{12}{c}}
\toprule
\multirow{2}{*}{Method}
& \multicolumn{4}{c}{Seen period (2804 trajectories)}
& \multicolumn{4}{c}{Regular unseen period (1655 trajectories)}
& \multicolumn{4}{c}{COVID-19 unseen period (1301 trajectories)} \\
\cmidrule(lr){2-5} \cmidrule(lr){6-9} \cmidrule(l){10-13}
& SD & SI & DARD & STVD
& SD & SI & DARD & STVD
& SD & SI & DARD & STVD \\
\midrule
LSTM
& 0.1489 & 0.2437 & 0.3872 & 0.6207
& 0.1740 & 0.2420 & 0.3810 & 0.6569
& 0.1493 & 0.2645 & 0.4358 & 0.6551 \\
MHSA
& 0.1897 & 0.1965 & 0.3578 & 0.5961
& 0.1715 & 0.2352 & 0.3762 & 0.6466
& 0.1582 & 0.2500 & 0.4182 & 0.6589 \\
DeepMove
& 0.2754 & 0.2267 & 0.3548 & 0.6510
& 0.2746 & 0.2257 & 0.3876 & 0.6714
& 0.2709 & 0.2338 & 0.3918 & 0.6704 \\
GetNext
& 0.0832 & 0.3220 & 0.2568 & 0.5752
& 0.0752 & 0.3215 & 0.2376 & 0.6408
& 0.0994 & 0.3351 & 0.2511 & 0.6438 \\
TrajGAIL
& 0.1737 & 0.0825 & 0.1287 & 0.5750
& 0.1954 & 0.0912 & 0.1578 & 0.6513
& 0.1915 & 0.0943 & 0.1427 & 0.6532 \\
DiffTraj
& 0.1977 & 0.0719 & 0.0932 & 0.6894
& 0.1944 & 0.0832 & 0.1283 & 0.6919
& 0.2159 & 0.0815 & 0.1405 & 0.6899 \\
GNPR-SID
& 0.1367 & 0.0706 & 0.1290 & 0.4673
& 0.1353 & 0.0626 & \underline{0.1272} & 0.5632
& 0.1481 & 0.0643 & \underline{0.1325} & 0.5925 \\
LLM4POI
& 0.0334 & 0.0699 & 0.1282 & 0.4435
& 0.0382 & 0.0763 & 0.1477 & 0.5682
& 0.0535 & 0.0859 & 0.1595 & 0.5663 \\
\midrule
LLMMove
& 0.0467 & 0.0602 & 0.1255 & 0.4674
& 0.0504 & 0.0663 & 0.1608 & 0.5834
& 0.0427 & 0.0737 & 0.1615 & 0.5831 \\
LLM-ZS
& \underline{0.0226} & 0.0709 & 0.1762 & 0.5005
& 0.0345 & 0.0909 & 0.2069 & 0.5759
& 0.0328 & 0.0875 & 0.1733 & 0.5910 \\
LLM-Mob
& 0.0734 & 0.0625 & 0.2746 & 0.4667
& 0.0715 & 0.0576 & 0.2992 & 0.5824
& 0.0821 & 0.0761 & 0.3141 & 0.5747 \\
LLMob
& 0.0583 & 0.0948 & \underline{0.0902} & 0.4756
& 0.0702 & 0.1130 & 0.1337 & 0.5766
& 0.0623 & 0.1243 & 0.1415 & 0.5865 \\
ELLMob
& 0.0227 & \underline{0.0366} & 0.1226 & \underline{0.4255}
& \underline{0.0260} & \underline{0.0383} & 0.1820 & \underline{0.5467}
& \underline{0.0264} & \underline{0.0408} & 0.1657 & \underline{0.5514} \\
\midrule
\textbf{MotiveMob}
& \textbf{0.0075} & \textbf{0.0251} & \textbf{0.0363} & \textbf{0.3872}
& \textbf{0.0143} & \textbf{0.0268} & \textbf{0.0696} & \textbf{0.5149}
& \textbf{0.0214} & \textbf{0.0385} & \textbf{0.0665} & \textbf{0.5219} \\
\bottomrule
\end{tabular}%
}
\vspace{2pt}
\end{table*}

Across all 24 combinations of user group, time period, and metric, MotiveMob consistently achieves the best performance. For seen users, it reduces the four divergences by an average of 36.3\% relative to the corresponding runner-up, with particularly large gains on SD, SI, and DARD. The improvements on STVD are smaller (0.7--2.8\%), but remain consistent across all three periods. For unseen users, MotiveMob retains the best results in every setting and yields an average relative error reduction of 31.1\% over the runner-up. Although its advantages on SI and STVD narrow under the COVID-19 shift, MotiveMob remains strongest overall, demonstrating robust generalization to both unseen users and temporal distribution shifts. Beyond user groups and temporal shifts, MotiveMob also performs consistently well across cities, achieving competitive results on both Tokyo and Osaka.

\FloatBarrier

\section{MotiveMob Results Across LLM Backbones}
\label{app:motivemob-backbone-comparison}

Table~\ref{tab:motivemob-backbone-comparison} shows that MotiveMob achieves
consistently strong results across different LLM backbones, indicating that its
effectiveness is not dependent on a particular backbone model.

\begin{table*}[!htbp]
\centering
\caption{MotiveMob results with different LLM backbones for seen and unseen
users across the three temporal settings. All values are JSD; lower is better.}
\label{tab:motivemob-backbone-comparison}
\setlength{\tabcolsep}{4pt}
\renewcommand{\arraystretch}{1.12}

\resizebox{\textwidth}{!}{%
\begin{tabular}{l*{12}{c}}
\toprule
\multirow{2}{*}{Backbone}
& \multicolumn{4}{c}{Seen period}
& \multicolumn{4}{c}{Regular unseen period}
& \multicolumn{4}{c}{COVID-19 unseen period} \\
\cmidrule(lr){2-5}
\cmidrule(lr){6-9}
\cmidrule(l){10-13}
& SD & SI & DARD & STVD
& SD & SI & DARD & STVD
& SD & SI & DARD & STVD \\
\midrule

\multicolumn{13}{l}{\textbf{Seen users}} \\
\addlinespace[2pt]

Qwen3.5-9B
& 0.0080 & 0.0088 & 0.0671 & 0.3552
& 0.0096 & 0.0116 & 0.0902 & 0.4464
& 0.0452 & 0.0173 & 0.1396 & 0.5728 \\

Gemma-4-E4B
& 0.0081 & 0.0191 & 0.0828 & 0.4041
& 0.0114 & 0.0192 & 0.0953 & 0.4731
& 0.0425 & 0.0200 & 0.1300 & 0.5431 \\

Gemma-2-9B
& 0.0099 & 0.0064 & 0.0641 & 0.3760
& 0.0097 & 0.0118 & 0.0889 & 0.4296
& 0.0482 & 0.0138 & 0.1319 & 0.5696 \\

\midrule

\multicolumn{13}{l}{\textbf{Unseen users}} \\
\addlinespace[2pt]

Qwen3.5-9B
& 0.0253 & 0.0268 & 0.0663 & 0.4798
& 0.0358 & 0.0288 & 0.0913 & 0.5411
& 0.0308 & 0.0278 & 0.1198 & 0.6431 \\

Gemma-4-E4B
& 0.0166 & 0.0373 & 0.0857 & 0.4913
& 0.0184 & 0.0382 & 0.1022 & 0.5514
& 0.0242 & 0.0299 & 0.1121 & 0.6267 \\

Gemma-2-9B
& 0.0362 & 0.0153 & 0.0727 & 0.4675
& 0.0366 & 0.0152 & 0.0918 & 0.5348
& 0.0244 & 0.0216 & 0.1184 & 0.6308 \\

\bottomrule
\end{tabular}%
}
\vspace{2pt}
\end{table*}
\FloatBarrier

\section{Analysis of Trajectory Termination}
\label{app:termination_analysis}

In MotiveMob, the motivation predictor $P_M$ has two related responsibilities. At each generation step, it either predicts a semantic motivation for the next transition or emits \texttt{END} to terminate the trajectory. The state predictor $P_S$ is invoked only when a non-terminal motivation is produced. In the direct ablation without $P_M$, a single predictor instead outputs either the next state or \texttt{END}. Thus, all configurations retain a learned termination decision, but differ in whether termination is separated from state generation through the motivation interface.

Since the timing of \texttt{END} determines the temporal span and number of generated visits, it may affect both SI and DARD. We therefore compare the trajectory lengths and termination behavior of the direct model and three variants sharing the same motivation predictor.

\begin{table*}[!htbp]
    \centering
    \caption{
    Trajectory-length and termination statistics for the motivation-mediation ablation. Mean length and termination rates are aggregated over all 4,671 test days from the six evaluation splits. SI and DARD are macro averages over the six splits. The maximum rollout length is 25 states.
    }
    \label{tab:termination_analysis}
    \resizebox{\textwidth}{!}{
    \begin{tabular}{l l c c c c}
        \toprule
        \textbf{State generation} &
        \textbf{\texttt{END} predictor} &
        \textbf{Mean length} &
        \textbf{Max-step rate} &
        \textbf{SI} $\downarrow$ &
        \textbf{DARD} $\downarrow$ \\
        \midrule
        Direct, w/o $P_M$ (IWM)
            & Direct predictor
            & 7.52 & 1.48\% & 0.0282 & 0.1401 \\

        $P_S$ (IWM)
            & $P_M$
            & 8.08 & 2.27\% & 0.0205 & \textbf{0.1087} \\

        $P_S$ (positive-only)
            & $P_M$
            & 7.42 & 2.06\% & \textbf{0.0176} & 0.1132 \\

        Rule-based state predictor
            & $P_M$
            & 9.06 & 1.76\% & 0.0215 & 0.1163 \\
        \midrule
        Ground truth
            & -- & 6.82 & -- & -- & -- \\
        \bottomrule
    \end{tabular}
    }
\end{table*}

Table~\ref{tab:termination_analysis} shows that the improvements in SI and DARD cannot be explained solely by matching the average trajectory length. The direct model generates 7.52 states per day on average, which is closer to
the ground-truth mean of 6.82 than the 8.08 states generated by the motivation-guided model with the IWM state predictor. Nevertheless, the direct model obtains substantially worse SI and DARD (0.0282 and 0.1401, respectively) than its motivation-guided counterpart (0.0205 and 0.1087). This result suggests that the semantic interface provided by $P_M$ contributes information beyond a simple control of trajectory length.

Reaching the rollout limit is also uncommon: only 1.48--2.27\% of trajectories reach the 25-state limit across the learned predictors. Consequently, differences in the number of capped trajectories are unlikely to be the main cause of the SI and DARD improvements. The same $P_M$ paired with different state predictors still produces noticeable metric variation. In particular, the positive-only state predictor achieves the best SI, whereas the IWM predictor achieves the best DARD. We therefore interpret motivation predictor as an important contributor of the observed improvements.
\FloatBarrier

\section{Trajectory-Level Effects of the Motivation Predictor}
\label{app:motivation-predictor-cases}

\begin{figure*}[!htbp]
  \centering
  \includegraphics[
    width=\textwidth,
    height=0.82\textheight,
    keepaspectratio
  ]{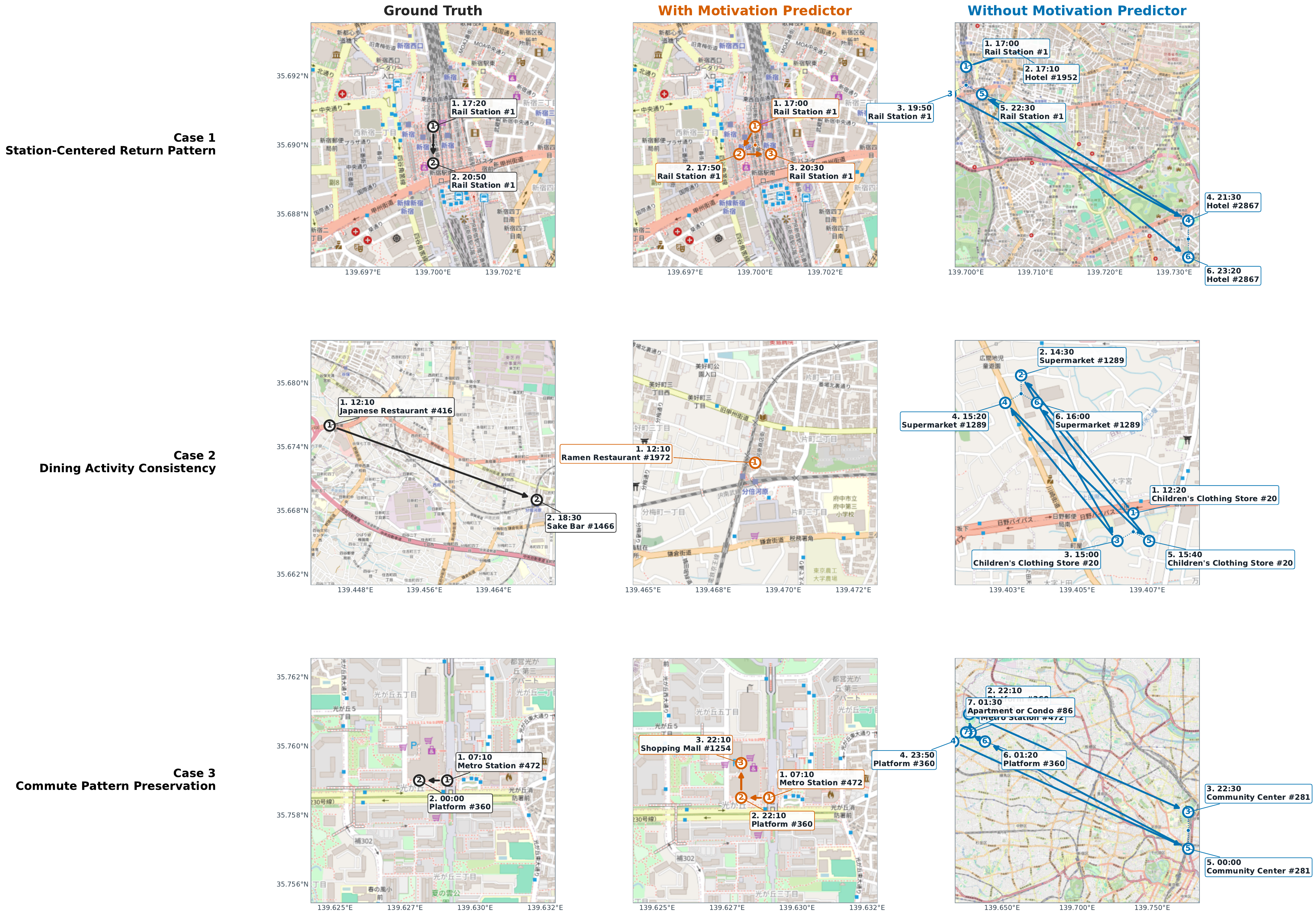}
  \caption{Trajectory comparisons with and without the trained motivation predictor. From left to right, the columns show the ground-truth trajectory, MotiveMob with the motivation predictor, and the corresponding ablation without it. Numbered markers indicate visit order, and labels show the visit time and POI.}
  \label{fig:motivation-predictor-cases}
\end{figure*}

Figure~\ref{fig:motivation-predictor-cases} provides three examples comparing MotiveMob with and without the trained motivation
predictor. The cases are selected from the same evaluation inputs used in the quantitative experiments. Both variants use the same trained state predictor, so the comparison isolates the contribution of motivation prediction. Lines connect consecutive visits and indicate visit order.

\paragraph{Case 1: Station-centered return pattern.}
The ground-truth trajectory revisits \textit{Rail Station\#1} from 17:20 to
20:50. With the motivation predictor, MotiveMob generates
\textit{Rail Station\#1} at 17:00, 17:50, and 20:30, retaining both the
station-centered activity and its evening timing. Without the predictor, the
trajectory becomes
\textit{Rail Station\#1} $\rightarrow$ \textit{Hotel\#1952}
$\rightarrow$ \textit{Rail Station\#1} $\rightarrow$
\textit{Hotel\#2867} $\rightarrow$ \textit{Rail Station\#1}
$\rightarrow$ \textit{Hotel\#2867}, introducing repeated hotel detours.

\paragraph{Case 2: Dining activity consistency.}
The reference trajectory moves from \textit{Japanese Restaurant\#416} at
12:10 to \textit{Sake Bar\#1466} at 18:30. MotiveMob with the predictor
generates \textit{Ramen Restaurant\#1972} at 12:10, preserving the dining
motivation and the time of the first activity, although it produces a shorter
trajectory. The ablation instead alternates between
\textit{Children's Clothing Store\#20} and \textit{Supermarket\#1289} six
times, shifting the trajectory from dining to retail and adding repetitive
visits.

\paragraph{Case 3: Commute pattern preservation.}
The ground truth contains \textit{Metro Station\#472} followed by
\textit{Platform\#360}. With the predictor, MotiveMob retains this ordered
transit pair and adds one nearby visit to \textit{Shopping Mall\#1254}.
Without the predictor, the sequence repeatedly moves between
\textit{Platform\#360} and the distant \textit{Community Center\#281} before
ending at \textit{Apartment or Condo\#86}, producing a longer and less
coherent chain.

Overall, these examples show that the motivation predictor helps preserve coherent activity timing and semantics, whereas removing it more often produces unrelated POIs, repetitive transitions, and implausible activity sequences. This pattern is consistent with the ablation results, where removing $P_M$ increases both SI and DARD across all six evaluation settings. Although some cases also exhibit fewer spatial detours with $P_M$, its primary contribution is improved temporal and activity-semantic coherence rather than a uniform improvement in spatial distribution.
\FloatBarrier

\section{Motivation Predictor Sensitivity to Early-Experience Supervision}
\label{app:motivation-predictor-sensitivity}

Following the early-experience training procedure \cite{zhang2025agent}, we investigate that whether one epoch of implicit world modeling (IWM) helps the subsequent motivation predictor.

We hold the state predictor fixed, using the IWM-trained LLM state engine in every experiment. We compare three initialization schemes for the motivation predictor: (1) direct imitation learning (IL) without IWM pretraining; (2) one epoch of IWM pretraining on a balanced mixture of observed and alternative transitions ($p_+=0.5$); and (3) one epoch of IWM pretraining on observed transitions only ($p_+=1.0$). For each initialization scheme, we subsequently train the motivation predictor for 4, 8, or 12 epochs. All other inference settings are held constant.

\begin{figure*}[t]
    \centering
    \includegraphics[
        width=\textwidth,
        height=0.90\textheight,
        keepaspectratio
    ]{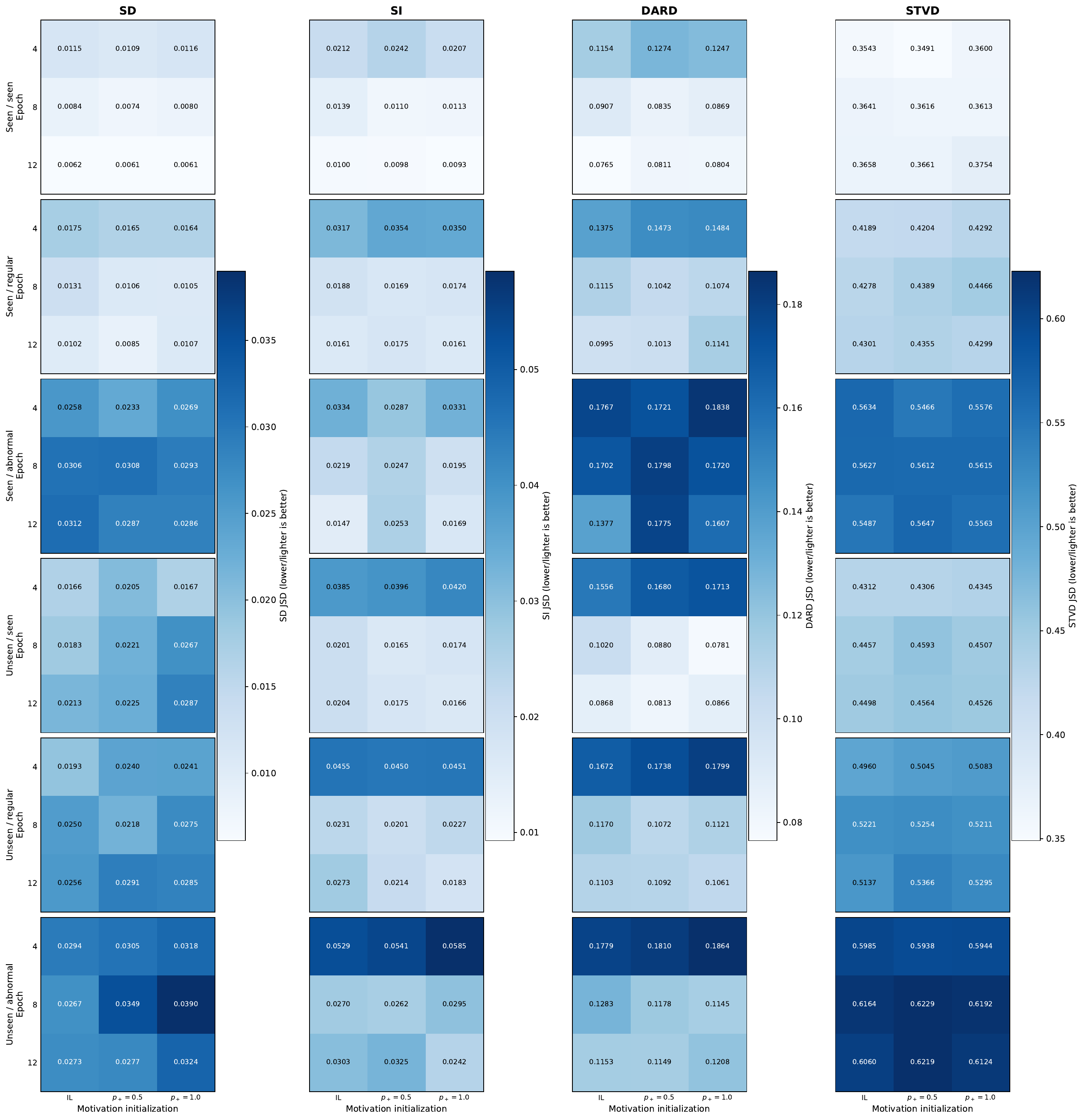}
    \caption{Motivation-predictor sensitivity to IWM initialization and training duration. Each evaluation split contains four $3\times3$ matrices corresponding to SD, SI, DARD, and STVD. Rows show results after 4, 8, and 12 epochs of motivation-predictor training, while columns show direct IL, balanced IWM initialization ($p_+=0.5$), and observed-only IWM initialization ($p_+=1.0$). Cell annotations report raw JSD values; lower values and lighter shading indicate better distributional alignment.}
    \label{fig:motivation-predictor-sensitivity}
\end{figure*}

Figure~\ref{fig:motivation-predictor-sensitivity} reports the four JSD metrics across the six evaluation splits. The horizontal axis indicates the motivation predictor initialization scheme, while the vertical axis indicates the number of motivation predictor training epochs. Each cell displays the corresponding raw JSD value.

\begin{table*}[t]
    \centering
    \caption{
        Mean JSD over the six evaluation splits. All metrics are lower-is-better. Mean rank is calculated by ranking the nine configurations separately for every test split and metric and
        then averaging the resulting 24 ranks.
    }
    \label{tab:motivation-predictor-sensitivity}
    \small
    \begin{tabular}{lrrrrrr}
        \toprule
        Initialization & Epoch & SD & SI & DARD & STVD & Mean rank \\
        \midrule
        IL
        & 4  & \textbf{0.0200} & 0.0372 & 0.1550 & 0.4771 & 5.42 \\
        IL
        & 8  & 0.0204 & 0.0208 & 0.1200 & 0.4898 & 5.04 \\
        IL
        & 12 & 0.0203 & 0.0198 & \textbf{0.1043} & 0.4857
        & \textbf{3.81} \\
        \midrule
        IWM, $p_+=0.5$
        & 4  & 0.0209 & 0.0378 & 0.1616 & \textbf{0.4742} & 5.42 \\
        IWM, $p_+=0.5$
        & 8  & 0.0213 & 0.0192 & 0.1134 & 0.4949 & 4.92 \\
        IWM, $p_+=0.5$
        & 12 & 0.0204 & 0.0207 & 0.1109 & 0.4969 & 5.02 \\
        \midrule
        IWM, $p_+=1.0$
        & 4  & 0.0213 & 0.0391 & 0.1658 & 0.4807 & 6.33 \\
        IWM, $p_+=1.0$
        & 8  & 0.0235 & 0.0196 & 0.1118 & 0.4934 & 4.92 \\
        IWM, $p_+=1.0$
        & 12 & 0.0225 & \textbf{0.0169} & 0.1115 & 0.4927 & 4.13 \\
        \bottomrule
    \end{tabular}
\end{table*}

The benefit of one epoch of IWM is most visible after 8 epochs of motivation predictor training. At this checkpoint, the $p_+=0.5$ and $p_+=1.0$ models outperform the epoch-matched IL baseline in 15 and 16 of the 24 test--metric comparisons, respectively. In particular, $p_+=0.5$ reduces the mean SI from $0.0208$ to $0.0192$ and the mean DARD from $0.1200$ to $0.1134$. However, its SD and STVD are slightly higher. Therefore, IWM initialization helps the motivation predictor learn SI and DARD more quickly, but does not improve every metric.

At epoch 12, the IWM advantage is no longer consistent. The selected IL model obtains the best overall mean rank ($3.81$) and the lowest mean DARD ($0.1043$), whereas the $p_+=1.0$ model obtains the lowest mean SI ($0.0169$). Thus, one epoch of IWM provides an intermediate-training advantage rather than a uniform improvement at the final checkpoint.
\FloatBarrier

\section{Inference-Time Validation and Termination Statistics}
\label{app:inference-time-validation-termination}

We report the operational statistics of MotiveMob's inference-time validation and termination mechanisms. Table~\ref{tab:validation-termination-statistics} reports the intervention statistics for the speed constraint. A \emph{step requiring resampling} is a transition for which at least one state proposal exceeds the speed limit. A step is considered \emph{resolved by resampling} if a valid proposal is obtained within the attempt budget. Otherwise, the same-category feasibility correction is invoked. We separately report whether the correction successfully identifies a feasible substitute.

We additionally apply repetition-aware termination after the generated trajectory reaches a minimum length of 15 states. Generation terminates when appending a candidate would extend a consecutively repeating location pattern. This operation preserves the generated prefix and is equivalent to emitting \texttt{END} before the repeated candidate.

\begin{table*}[!htbp]
\centering
\caption{Inference-time validation and termination statistics. All percentages are normalized by generated states, except repetition termination, which is normalized by generated trajectories.}
\label{tab:validation-termination-statistics}
\setlength{\tabcolsep}{3.2pt}
\renewcommand{\arraystretch}{1.12}
\scriptsize
\resizebox{\textwidth}{!}{%
\begin{tabular}{llrrrrrrrr}
\toprule
User
& Period
& Days
& States
& Resampling
& Resampling
& Correction
& Feasibility
& No feasible
& Repetition \\
status
&
&
&
&
required
& resolved
& invoked
& corrected
& substitute
& terminated \\
\midrule
Seen & Seen
& 908
& 7,647
& 932 (12.2\%)
& 528 (6.9\%)
& 404 (5.3\%)
& 231 (3.0\%)
& 173 (2.3\%)
& 121 (13.3\%) \\

Seen & Regular unseen
& 690
& 5,446
& 686 (12.6\%)
& 415 (7.6\%)
& 271 (5.0\%)
& 134 (2.5\%)
& 137 (2.5\%)
& 88 (12.8\%) \\

Seen & COVID-19 unseen
& 392
& 3,218
& 830 (25.8\%)
& 459 (14.3\%)
& 371 (11.5\%)
& 237 (7.4\%)
& 134 (4.2\%)
& 77 (19.6\%) \\
\midrule
Unseen & Seen
& 1,345
& 10,066
& 2,582 (25.7\%)
& 1,534 (15.2\%)
& 1,048 (10.4\%)
& 629 (6.2\%)
& 419 (4.2\%)
& 207 (15.4\%) \\

Unseen & Regular unseen
& 795
& 6,016
& 1,610 (26.8\%)
& 929 (15.4\%)
& 681 (11.3\%)
& 432 (7.2\%)
& 249 (4.1\%)
& 150 (18.9\%) \\

Unseen & COVID-19 unseen
& 541
& 4,294
& 1,224 (28.5\%)
& 716 (16.7\%)
& 508 (11.8\%)
& 260 (6.1\%)
& 248 (5.8\%)
& 97 (17.9\%) \\
\midrule
\multicolumn{2}{l}{\textbf{All splits}}
& \textbf{4,671}
& \textbf{36,687}
& \textbf{7,864 (21.4\%)}
& \textbf{4,581 (12.5\%)}
& \textbf{3,283 (8.9\%)}
& \textbf{1,923 (5.2\%)}
& \textbf{1,360 (3.7\%)}
& \textbf{740 (15.8\%)} \\
\bottomrule
\end{tabular}%
}
\end{table*}
\FloatBarrier

\section{Additional Spatial Distribution Alignment Results}
\label{app:additional-spatial-distributions}

\begin{figure*}[!t]
  \centering
  \includegraphics[
    width=\textwidth,
    height=0.78\textheight,
    keepaspectratio
  ]{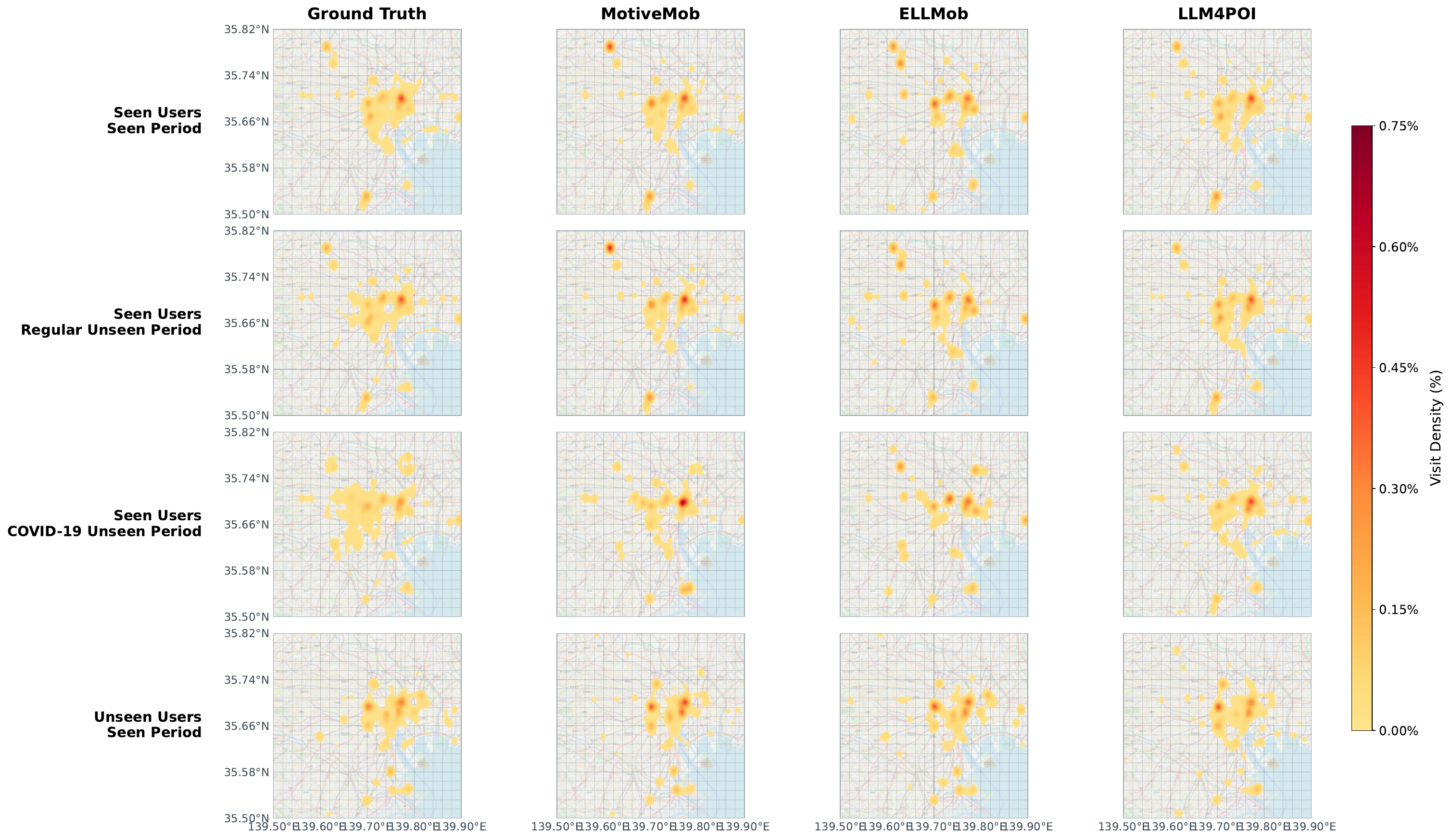}
  \caption{Ground-truth and predicted spatial mobility distributions for four additional evaluation settings. From top to bottom, the rows show seen users in the seen period, seen users in the regular unseen period, seen users in the COVID-19 unseen period, and unseen users in the seen period. Darker red indicates higher visit density.}
  \label{fig:additional-spatial-alignment}
\end{figure*}

Figure~\ref{fig:additional-spatial-alignment} presents the four evaluation settings omitted from the main spatial-distribution analysis. All three methods broadly reproduce the major ground-truth activity centers, with relatively modest differences when users and temporal patterns are familiar. Compared with LLM4POI, MotiveMob generally forms more compact hotspots with less peripheral dispersion. MotiveMob more consistently preserves the allocation of visit mass across the main activity regions, particularly under the COVID-19 shift. Overall, these results support the main finding that MotiveMob maintains robust spatial alignment as the user or temporal setting changes.

\FloatBarrier

\end{document}